\documentclass{ieeeaccess}
\usepackage{cite}
\usepackage{amsmath,amssymb,amsfonts}
\usepackage{algorithm}
\usepackage{algorithmicx}
\usepackage{algpseudocode}
\usepackage{qcircuit}
\usepackage{braket}
\usepackage{hyperref}       
\usepackage{url}            
\usepackage{graphicx}
\usepackage{textcomp}
\usepackage{caption}
\usepackage{subcaption}
\usepackage{tablefootnote}
\usepackage{xcolor}
\usepackage{float}
\usepackage{soul}

\usepackage[flushleft]{threeparttable}

\definecolor{amber}{rgb}{1.0, 0.75, 0.0}

\renewcommand{\textcolor}[2]{{\color{black}#2}}

\renewcommand{\figureautorefname}{Fig.~\negthinspace}
\renewcommand{\tableautorefname}{Table \negthinspace}
\renewcommand{\sectionautorefname}{Sec.~\negthinspace}

\DeclareMathOperator*{\E}{\mathbb{E}}
\def\BibTeX{{\rm B\kern-.05em{\sc i\kern-.025em b}\kern-.08em
    T\kern-.1667em\lower.7ex\hbox{E}\kern-.125emX}}
\begin{document}
\history{Date of publication xxxx 00, 0000, date of current version xxxx 00, 0000.}
\doi{10.1109/ACCESS.2017.DOI}

\title{Variational Quantum Circuits for Deep Reinforcement Learning}
\author{
\uppercase{Samuel Yen-Chi Chen}\authorrefmark{1}, 
\uppercase{Chao-Han Huck Yang}\authorrefmark{2}, 
\uppercase{Jun Qi}\authorrefmark{2},
\uppercase{Pin-Yu Chen}\authorrefmark{3},
\uppercase{Xiaoli Ma}\authorrefmark{2},
\IEEEmembership{Fellow, IEEE},
\uppercase{Hsi-Sheng Goan}\authorrefmark{1,4},
}
\address[1]{Department of Physics and
  Center for Theoretical Physics,
  National Taiwan University,
  Taipei, Taiwan}
\address[2]{School of Electrical and Computer Engineering,
  Georgia Institute of Technology,
  Atlanta, GA, USA}
\address[3]{IBM Research
  Yorktown Heights, NY, USA}
\address[4]{
  Center for Quantum Science and Engineering
  National Taiwan University,
  Taipei, Taiwan}

\tfootnote{This work was supported in part by the Ministry of Science and Technology (MOST)
of Taiwan under Grants No.~MOST 106-2112-M-002-013-MY3, No.~MOST 108-2627-E-002-001, No.~MOST 108-2622-8-002-016,
and No.~MOST 107-2627-E-002-001-MY3, and by the National
Taiwan University under Grants No.~NTU-CC-108L893202, and No.~NTU-CC-109L892002.}

\markboth
{S. Y.-C. Chen \headeretal: Variational Quantum Circuits for Deep Reinforcement Learning}
{S. Y.-C. Chen \headeretal: Variational Quantum Circuits for Deep Reinforcement Learning}

\corresp{Corresponding authors: Samuel Yen-Chi Chen (e-mail: ycchen1989@gmail.com), Hsi-Sheng Goan (e-mail: goan@phys.ntu.edu.tw).}

\begin{abstract}
  The state-of-the-art machine learning approaches are based on classical von Neumann computing architectures and have been widely used in many industrial and academic domains. With the recent development of quantum computing, researchers and tech-giants have attempted new quantum circuits for machine learning tasks. However, the existing quantum
computing platforms are hard to simulate classical deep learning models or problems because of the intractability of deep quantum circuits. Thus, it is necessary to design feasible quantum algorithms for quantum machine learning for noisy intermediate scale quantum (NISQ) devices. This work explores variational quantum circuits for deep reinforcement learning. Specifically, we reshape classical deep reinforcement learning algorithms like experience replay and target network into a representation of variational quantum circuits.
Moreover, we use a quantum information encoding scheme to reduce the number of model parameters compared to classical neural networks.
To the best of our knowledge, this work is the first proof-of-principle demonstration of variational quantum circuits to approximate the deep $Q$-value function for
decision-making and policy-selection reinforcement learning 
with experience replay and target network.
Besides, our variational quantum circuits can be deployed in many near-term 
NISQ machines. 
\end{abstract}


\titlepgskip=-15pt

\maketitle

\section{Introduction}
\label{sec:introduction}
Deep Learning (DL)~\cite{lecun2015deep} has been widely used in many machine learning domains, such as computer vision~\cite{Simonyan2014VeryRecognition, Szegedy2014GoingConvolutions, Voulodimos2018DeepReview}, natural language processing~\cite{Sutskever2014SequenceNetworks}, communication network congestion control~\cite{kao2019reinforcement}, and mastering the game of Go~\cite{Silver2016MasteringSearch}. The successful deployment of DL is primarily attributed to the improvement of new computer architectures associated with powerful computing capabilities in the past decades. Many researchers also utilized DL-based data analysis methods on fundamental 
physics researches such as quantum many-body physics\cite{Borin2019ApproximatingStates,Carleo2019NetKet:Systems,Carleo2019MachineSciences}, phase-transitions~\cite{Canabarro2019UnveilingLearning}, quantum control~\cite{An2019DeepControl,Flurin2018UsingObservations}, and quantum error correction~\cite{Andreasson2018QuantumLearning,Nautrup2018OptimizingLearning}. In the meantime, great efforts from both the physics and machine learning community have dedicated to and empowered quantum computation. Quantum computing machines have been brought to the market 
(e.g., IBM's and D-Wave's hardware solutions~\cite{cross2018ibm, lanting2014entanglement}), but a large-scale quantum circuits cannot be faithfully employed upon the quantum computing platforms due to the lack of quantum error correction~\cite{gottesman1997stabilizer,gottesman1998theory}. Therefore, Mitarai et al.
design approximate quantum algorithms, circuits and encoding schemes~\cite{Mitarai2018QuantumLearning} on the devices with noise tolerance. 
More specifically, the work takes the advantages of quantum entanglement~\cite{Mitarai2018QuantumLearning,du2018expressive} in quantum computing to reduce the model size into an essentially small number and take advantage of 
the iterative optimization to reduce 
the quantum circuit depth to a practically low value such that hybrid quantum-classical  algorithms can be realized on the available quantum platforms which are named as noisy intermediate-scale quantum (NISQ) machines~\cite{Preskill2018quantumcomputingin,du2018expressive}. 

By taking the strengths of quantum computing
with significantly fewer parameters~\cite{du2018expressive},
variational quantum circuits on NISQ have succeeded in implementing standard classification and clustering algorithms on classical benchmark datasets~\cite{Mitarai2018QuantumLearning,Schuld2018Circuit-centricClassifiers,havlivcek2019supervised}.
Besides, it is also possible to employ quantum circuits for implementing new DL algorithms like generative adversarial networks \cite{goodfellow2014generative} (GAN) on NISQ machines.
These frameworks and development pave the way \textcolor{black}{towards} applications of near-term quantum
devices for quantum machine learning.
However, to the best of our knowledge, variational circuits on current NISQ computing for deep neural network based decision making and policy selection problems have not been discussed, which constrains the application of NISQ in many machine learning scenarios with \textcolor{black}{sequential decision making.}



Since reinforcement learning (RL) and deep reinforcement learning (DRL) are two paradigms of complex sequential decision-making systems
and satisfy the requirements of automatic policy learning under uncertainty, 
our work focuses on the empowerment of DRL on NISQ computation, which refers to an agent interacting with the environment to gain knowledge of backgrounds and deriving the policy of decision making accordingly~\cite{SuttonReinforcementIntroduction, Mnih2016AsynchronousLearning}. We propose a novel variational quantum circuit feasible on the current NISQ platform hybridized with iterative parameter optimization on a classical computer to
resolve the circuit-depth challenges.
Furthermore, we generalize variational quantum circuits to standard DRL based \emph{action-value} function approximation~\cite{Mnih2016AsynchronousLearning,Mnih2015Human-levelLearning}. Finally, we analyze the policy reward and the memory cost for performance of our variational quantum circuits (VQC) based DRL in comparison with standard RL and DRL approaches in the context of \emph{frozen-lake}~\cite{Brockman2016OpenAIGym} and \emph{cognitive-radio}~\cite{gawlowicz2018ns3} environments. The frozen lake is a simple maze environment in openAI Gym \cite{Brockman2016OpenAIGym} and is a typical and simple example that is demonstrated in standard RL. Cognitive radio is a wireless technology that enables for optimizing the use of available communication channels between users and has been studied by the standard 
machine learning technique~\cite{gawlowicz2018ns3, kao2019reinforcement}.
Under current limitations on the scale of quantum machines and the capabilities of quantum simulations, we select the frozen-lake and cognitive-radio environments
for the proof-of-principle quantum machine learning study.
To the best of our knowledge, this work is the first demonstration of variational quantum circuits to the DRL-based 
decision-making and policy-selection problems.

\begin{figure}[ht]

\center
\includegraphics[width=1.0\linewidth]{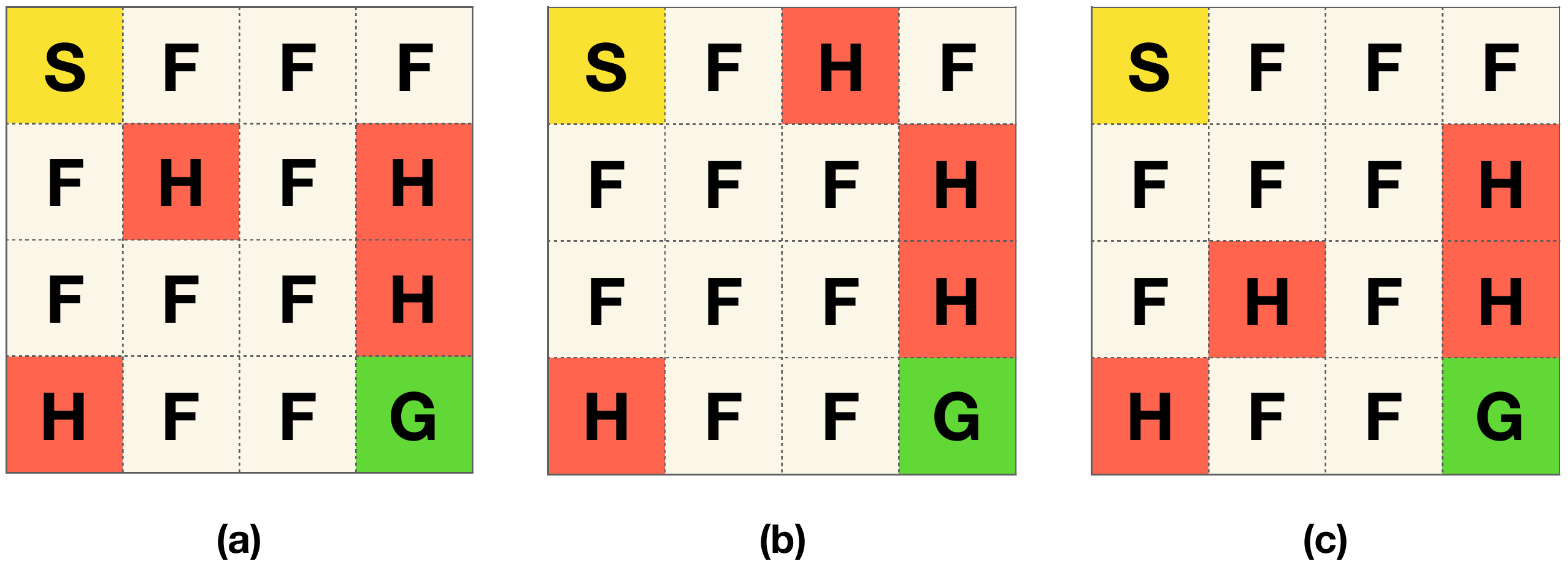}

\caption[Environment: Frozen Lake]{{\bfseries Frozen-Lake environment for the variational quantum DRL agent.}
In this frozen-lake environment, the RL agent is expected to go from the start location (S) to the goal location (G). There are several holes (H) on the way, and the agent should learn to avoid stepping into these hole locations. Furthermore, we set a negative reward for each step the agent takes. The agent is expected to learn the policy that going from S to G with the shortest path possible. \textcolor{black}{In this work, we train the agents on three configurations of the frozen-lake environment shown in (a), (b) and (c) separately.}
 }
\label{VQDQN_Env_FrozenLake}
\end{figure}

\begin{figure}[ht]
\center
\includegraphics[width=0.9\linewidth]{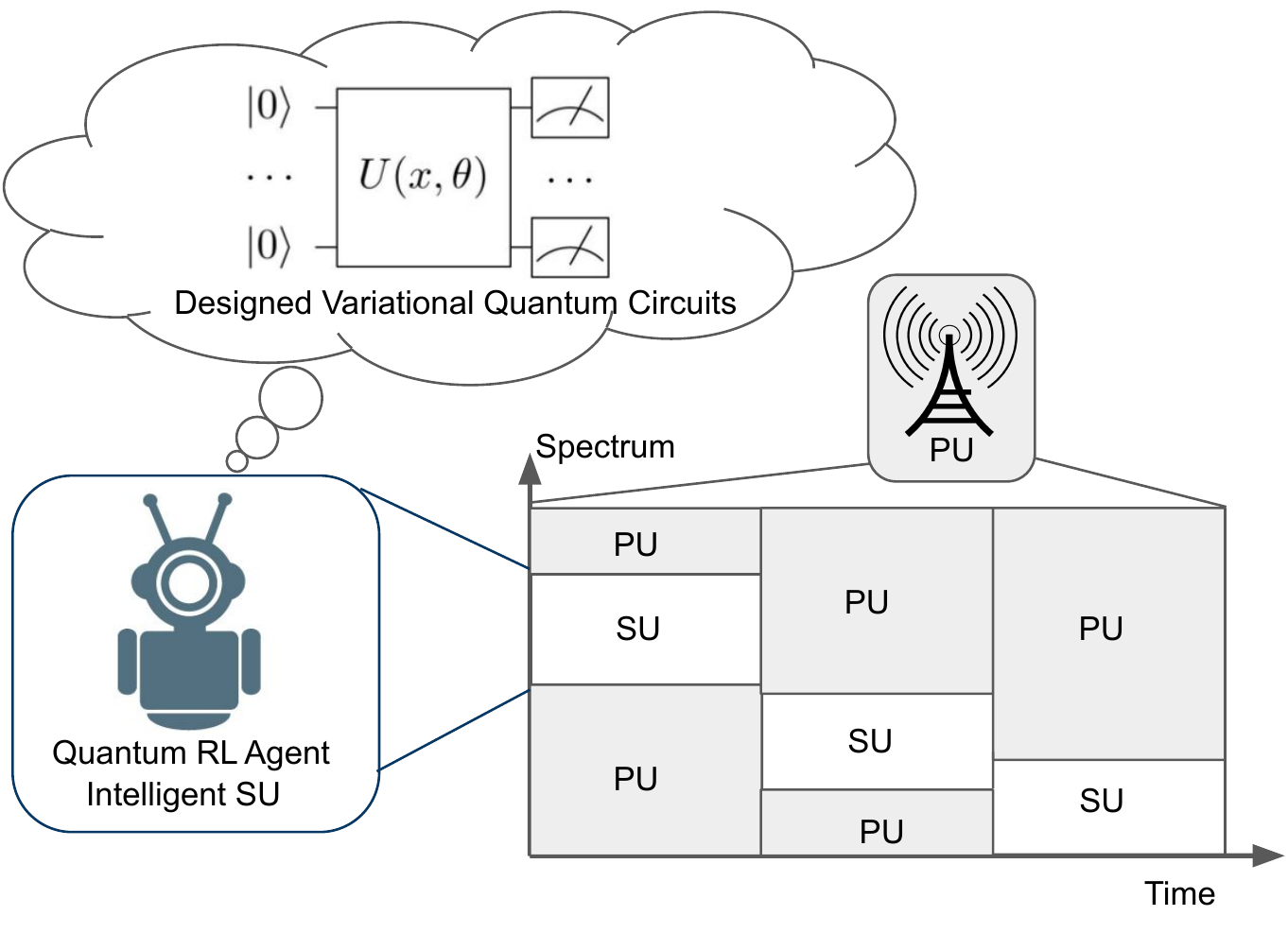}
\caption[Environment: Cognitive Radio]{{\bfseries Cognitive-Radio environment for the variational quantum DRL agent.}
\textcolor{black}{In the cognitive-radio environment, the agent is expected to select a channel that is free of interference in each time step. For example, there is a \emph{primary user} (PU) that will occupy a specific channel in each time step periodically. Our agent, which is the \emph{secondary user} (SU), can only select the channels that are not occupied without the knowledge of the PU in advance. The agent is expected to learn the policy through the interaction with the environment. }}
\label{VQDQN_Env_CognitiveRadio}
\end{figure}

\begin{figure}[ht]
\center
\includegraphics[width=0.7\linewidth]{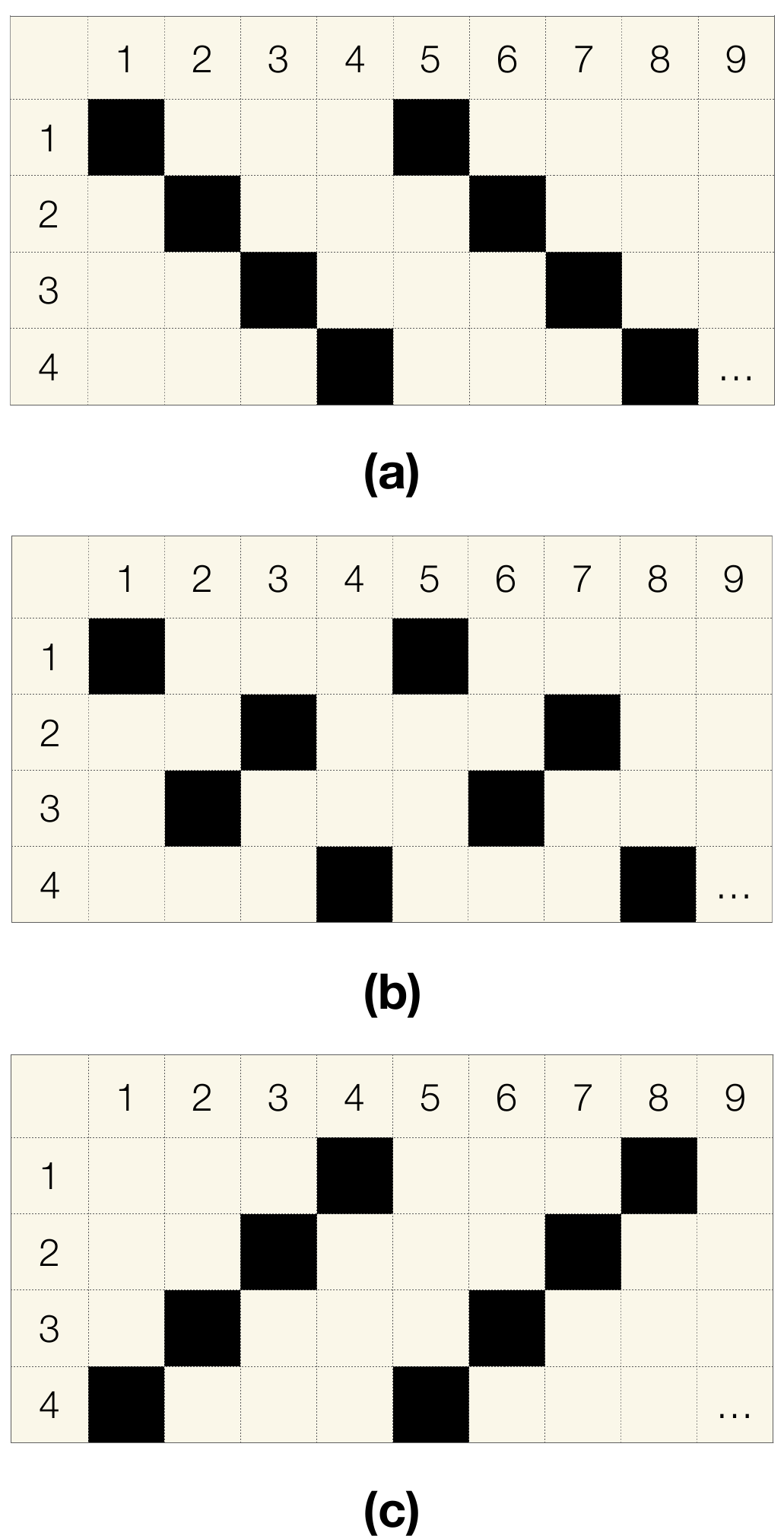}
\caption[Environment: Cognitive Radio 3 configs]{{\bfseries Cognitive-Radio environment with periodical channel-changing pattern
for the variational quantum DRL agent.}
\textcolor{black}{We provide three configurations for the cognitive-radio experiment; the first setting (a) is the main configuration for experiments on a different number of channels and experiments in noisy situations. The other two configurations in (b) and (c) are only tested in the case of 4 channels, the purpose of these additional experiments is to demonstrate that the proposed framework is generally applicable in different scenarios. } }
\label{VQDQN_Env_CognitiveRadio_3_configs}
\end{figure}

\section{Reinforcement Learning}
\emph{Reinforcement learning} is a machine learning paradigm in which an \emph{agent} interacts with an \emph{environment} $\mathcal{E}$ over a number of discrete time steps~\cite{SuttonReinforcementIntroduction}. At each time step $t$, the agent receives a \emph{state} or \emph{observation} $s_t$ and then chooses an \emph{action} $a_t$ from a set of possible actions $\mathcal{A}$ according to its \emph{policy} $\pi$. The policy is a function mapping the state $s_t$ to action $a_t$. In general, the policy can be stochastic, which means that given a state $s$, the action output can be a probability distribution. After executing the action $a_t$, the agent receives the state of the next time step $s_{t+1}$ and a scalar \emph{reward} $r_t$. The process continues until the agent reaches the terminal state. An \emph{episode} is defined as an agent starting from a randomly selected initial state and following the aforementioned process all the way through the terminal state.

Define $R_t = \sum_{t'=t}^{T} \gamma^{t'-t} r_{t'}$ as the total discounted return from time step $t$, where $\gamma$ is the discount factor that lies in $(0,1]$. In principle, $\gamma$ is provided by the investigator to control how future rewards are given to the decision making function. When a large $\gamma$ is considered, the agent takes into account future rewards no matter what a discount rate is. As to a small $\gamma$, an agent can quickly ignore future rewards within a few time steps. 
The goal of the agent is to maximize the expected return from each state $s_t$ in the training process. The \emph{action-value function} or \emph{Q-value function} $Q^\pi (s,a) = \mathbb{E}[R_t|s_t = s, a]$ is the expected return for selecting an action $a$ in state $s$ based on policy $\pi$. The optimal action value function $Q^*(s,a) = \max_{\pi} Q^\pi(s,a)$ gives a maximal action-value across all possible policies. The value of state $s$ under policy $\pi$, $V^\pi(s) = \mathbb{E}\left[R_t|s_t = s\right]$, is the agent's expected return by following policy $\pi$ from the state $s$. \textcolor{black}{The classical temporal difference (TD) error~\cite{SuttonReinforcementIntroduction} is used to update value function in reinforcement learning tasks.}




\subsection{Q-Learning}
Q-learning \cite{SuttonReinforcementIntroduction} is a model-free RL algorithm. Before the learning process begins, $Q$ is initially assigned to an arbitrary fixed value (chosen by the programmer). Then, at each time, the agent selects an action $a_t$ (using, e.g., $\epsilon$-greedy policy derived from $Q$), observes a reward $r_t$, and enters a new state $s_{t+1}$ (that may depend on both the previous state $s_t$ and the selected action), and then $Q$ is updated with the learning rate $\alpha$. The Q-learning is an \emph{off-policy} learner since it updates its Q-value
using the observed reward $r_t$ and the \textcolor{black}{maximum reward $\max _{a} Q\left(s_{t+1}, a\right)$ for the next state $s_{t+1}$ over all possible actions $a$.} The updating is done according to the benchmark formula:\footnote{The formula and loss are from the original DQN work Mnih et. al. ~\cite{Mnih2015Human-levelLearning}. }
\begin{align}
  Q\left(s_{t}, a_{t}\right) \leftarrow & \, Q\left(s_{t}, a_{t}\right)\nonumber\\
  &+\alpha\left[r_{t}+\gamma \max _{a} Q\left(s_{t+1}, a\right)-Q\left(s_{t}, a_{t}\right)\right].
\end{align}

\subsection{State-Action-Reward-State-Action (SARSA)} 
An SARSA~\cite{SuttonReinforcementIntroduction} agent interacts with the environment and updates the policy based on the undertaking actions.
\textcolor{black}{The $Q$ value represents the possible reward received in the next time step for taking action $a_{t}$ in state $s_{t}$, plus the discounted future reward received from the next state-action observation,
and is updated
by temporal difference with transitions from state-action pair $(s_{t}, a_{t})$ to state-action pair $(s_{t+1}, a_{t+1}$),
adjusted by the learning rate $\alpha$ as:
~\footnote{We follow the classical SARSA definition from Sutton et. al.~\cite{SuttonReinforcementIntroduction}. }}
\begin{align}
  Q\left(s_{t}, a_{t}\right) \leftarrow & \, Q\left(s_{t}, a_{t}\right)\nonumber \\
  &+\alpha\left[r_{t}+\gamma Q\left(s_{t+1}, a_{t+1}\right)-Q\left(s_{t}, a_{t}\right)\right].
\end{align}



\subsection{Deep Q-Learning}
The action-value function $Q(s, a)$ can be explicitly represented by a two-dimensional table with a total number of entries $s \times a$, that is, the number of possible states times the number of possible actions. However, when the state space or the action space is large or even continuous the tabular method is unfeasible. In such a situation, the action-value function is represented with function approximators such as neural networks~\cite{Mnih2016AsynchronousLearning, Mnih2015Human-levelLearning}. This neural-networks-based reinforcement learning is called \emph{deep reinforcement learning} (DRL). 

\textcolor{black}{
The employment of neural networks for function approximators to represent the $Q$-value function has been studied extensively~\cite{Mnih2016AsynchronousLearning, Mnih2015Human-levelLearning} and succeeded in many tasks like playing video games. In this setting, the action-value function $Q(s, a;\theta)$ is parameterized by $\theta$, which can be derived by a series of iterations from a variety of optimization methods adopted from other machine learning tasks. The simplest form is the $\emph{Q-learning}$. In this method, the goal is to directly approximate the optimal action-value function $Q^*(s, a)$ by minimizing the mean square error (MSE) loss function: 
\begin{equation}
L(\theta) = \mathbb{E}[(r_t + \gamma  \max_{a'} Q(s_{t+1},a';\theta^-) - Q(s_t,a_t;\theta))^2].
\end{equation}
Here, the prediction is $Q(s_t,a_t;\theta)$,
where $\theta$ is the parameter of the policy network,
and the target is $r_t + \gamma  \max_{a'} Q(s_{t+1},a';\theta^-)$, where $\theta^-$ is the parameter of the target network and 
$s_{t+1}$ is the state encountered after playing action $a_t$ 
at state $s_t$.
The loss function in DRL is normally hard to converge and is likely to get divergent when a nonlinear approximator like a neural network is used to represent the action-value function~\cite{Mnih2015Human-levelLearning}. There are several possible culprits. When the states or observations are serially correlated with each other along the trajectory, thereby violating the assumption that the sample needs to be independent and identically distributed (IID), the $Q$ function changes dramatically and changes the policy at a relatively large scale. In addition, the correlation between the action-value $Q$ and the target values $r_t + \gamma  \max_{a'} Q(s_{t+1},a')$ can be large. Unlike the supervised learning where the targets are given and invariant, the setting of DRL
allows targets to vary with $Q(s,a)$, causing $Q(s,a)$ to chase a nonstationary target.}

The \emph{deep Q-learning} (DQL) or \emph{deep Q-network} (DQN) presented in the work \cite{Mnih2015Human-levelLearning} addressed these issues through two mechanisms:
\begin{itemize}
\item \emph{Experience replay}: To perform experience replay, one stores each transition the agent encounters. The transition is stored as a tuple in the following form:
$(s_t, a_t, r_t, s_{t+1})$ at each time step $t$. To update the $Q$-learning parameters,
one randomly samples a batch of experiences from the replay memory and then performs gradient descent with the following MSE loss function:
$L(\theta) = \E[(r_t + \gamma  \max_{a'} Q(s_{t+1},a';\theta^-) - Q(s_t,a_t;\theta))^2]$, where the loss function is calculated over the batch sampled from the replay memory. The key importance of experience replay is to lower the correlation of inputs for training the $Q$-function.
\item \emph{Target Network}: $\theta^-$ is the parameter of the target network and these parameters are only updated at every finite time steps. This setting helps to stabilize the $Q$-value function training since the target is relatively stationary compared to the action-value function.
\end{itemize}



\section{Testing Environments}
\label{sec:test_env}
To study the performance of a reinforcement learning agent, we need to specify the \emph{environment} for the test. We will consider the
\emph{frozen-lake}~\cite{Brockman2016OpenAIGym} and \emph{cognitive-radio}~\cite{gawlowicz2018ns3} environments.
The reason why we choose the frozen-lake environment is two-fold. First is that it is a fairly simple and commonly tested example in standard RL, and if the dimension of the problem size is not too large, the simulation is feasible with available quantum simulators and NISQ devices, and the time consumption for the experiment is reasonable too. The second is that we want to demonstrate that quantum circuits are capable of learning sequential decision making process (also called \emph{policy}). The choice of cognitive-radio environment is that we want to demonstrate some kinds of real-world applications and
the complexity of this environment is comparable to the frozen-lake environment.

\subsection{Frozen Lake}
The first testing environment we consider in this work is the frozen lake, a simple maze environment in openAI Gym~\cite{Brockman2016OpenAIGym}. In this environment, the agent standing on a frozen lake is expected to go from the start location (S) to the goal location (G) (see~\figureautorefname{\ref{VQDQN_Env_FrozenLake}}).
Since the lake is not all frozen, there are several holes (H's) on the way, and the agent should learn to avoid stepping into these hole locations, otherwise the agent will get a large negative reward and the episode will terminate. Furthermore, the agent is also expected to take the shortest possible path. In order to accomplish this, we set a little negative reward on each move.
\textcolor{black}{Here we demonstrate three different configurations of the frozen-lake environment, as shown in~\figureautorefname{\ref{VQDQN_Env_FrozenLake}}, for the training.}

\textbf{The frozen-lake environment mapping is:}
\begin{itemize}
\item Observation: observed records of all time steps.
\item Action: there are four actions {LEFT, DOWN, RIGHT, UP} in the action space. How to choose the action in a variational quantum circuit will be described in Sec.~\ref{sec:Env-setup}. 
\item Reward: The rewards in this environment are $+1.0$ for successfully achieving the goal, $-0.2$ for failing the task, which is stepping into one of the holes. Moreover, to encourage the agent to take the shortest path, there is also a $-0.01$ reward for each step taken. 
\end{itemize} 

\subsection{Cognitive Radio}
In the second testing environment we study the proposed variational quantum-DQN or -DQL (VQ-DQN; VQ-DQL) agent in a real-world application. We consider the cognitive-radio experiment. In this setting, the agent is expected to select a channel
\textcolor{black}{that is not occupied or interfered by a primary user} (see~\figureautorefname{\ref{VQDQN_Env_CognitiveRadio}}).
If the agent succeeds, then it will get $+1$ reward, otherwise it will get $-1$ reward. Note that the episode will terminate if the agent collects three failed selections or the agent plays more than 100 steps.
This task is crucial for the modern wireless multi-channel environment since channels are possibly occupied or under interference.
For the demonstration in this work,  we consider that
  there are $n$ possible channels for the agent to select and the channel-changing by the primary user follows a simple periodic pattern with $n$ time-steps in a full cycle. Three different configurations of the cognitive-radio environment  in the case of four channels for the training considered here are illustrated
  in~\figureautorefname{\ref{VQDQN_Env_CognitiveRadio_3_configs}}.

\textbf{The cognitive-radio environment mapping is:}
\begin{itemize}
\item Observation: $ns3$~\cite{gawlowicz2018ns3} statistics with the radio channels capacity, with a customized channel number = $n$. (e.g., a state of \textbf{[1 0 0 0]} represents for $n=4$ channels and a primary user on the $1$st channel.) 

\item Action: selecting one channel for the secondary user accessing a radio channel out of $n$ channels. How to choose the action in a variational quantum circuit for the cognitive-radio scenario will be described in Sec.~\ref{sec:Env-setup}.

\item Reward: $-1$ for a collision with the primary user; $+1$ for no collision. \textcolor{black}{The list of score presenting rewards in the testing environments is shown in Table~\ref{reward_list}. Agent can achieve a maximum score of 100. }
\end{itemize}

\begin{table}[ht]
    \caption{List of rewards in our frozen-lake
    and cognitive-radio testing environments. In the frozen lake, the environment is non-slippery.
This setting encourages the agent not only to achieve the goal but also to select the shortest path.} \label{reward_list}
    \begin{subtable}{.5\linewidth}
      \centering
        \caption{Frozen-Lake}
        \begin{tabular}{c|c}
            Location & Reward \\ \hline \hline
            HOLE & -0.2 \\
            GOAL & +1.0 \\
            OTHER & -0.01 \\
        \end{tabular}
    \end{subtable}%
    \begin{subtable}{.5\linewidth}
      \centering
        \caption{Cognitive-Radio}
        \begin{tabular}{c|c}
            Location & Reward \\ \hline \hline
            Occupied Channel & -1.0 \\
            Available Channel & +1.0 \\
        \end{tabular}
    \end{subtable}
\end{table}
 \subsection{Variational Quantum Deep Q-Learning }

\begin{figure}[htbp]
\center
\includegraphics[width=1.\linewidth]{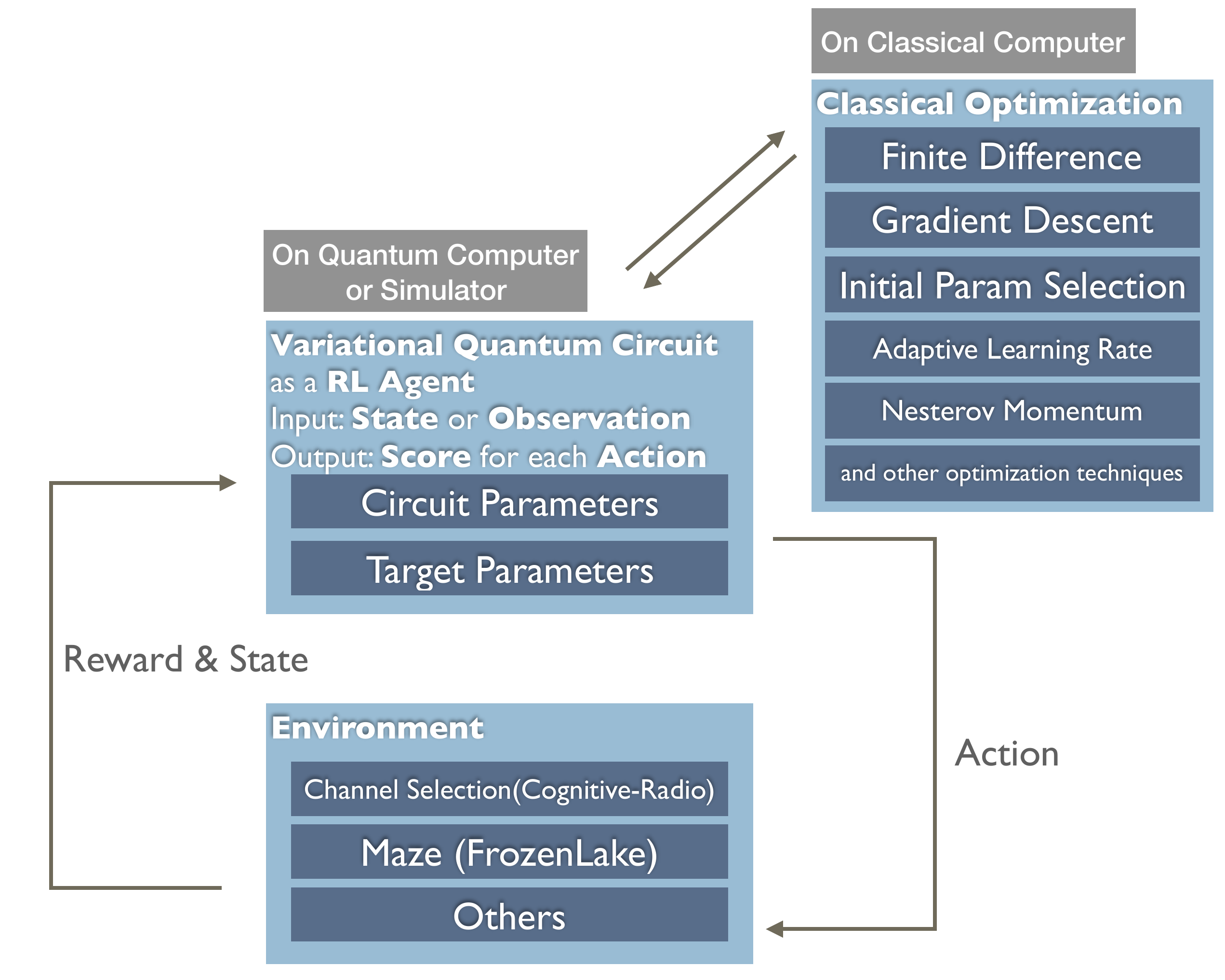}
\caption[Overview: Variational Quantum Circuit for Deep Q Learning (DQL)]{{\bfseries Overview of variational quantum circuits for DRL.}
In this work, we study the capability of variational quantum circuits in performing DRL tasks. This DRL agent includes a quantum part and a classical part. Under current limitations on the scale of quantum machines and the capabilities of quantum simulations, we select frozen-lake and cognitive-radio environments for the proof-of-principle study. The proposed framework is rather general and is expected to solve complicated tasks when larger-scale quantum machines are available.
 }
\label{VQDQN_Overview}
\end{figure}
\section{Variational Quantum Circuits and Deep Q-Learning}
\label{sec:vqc}
The variational quantum circuit is a hybrid quantum-classical approach which leverages the strengths of quantum and classical computation. It is one type of quantum circuits with tunable parameters which
are optimized in an iterative manner by a classical computer. These parameters can be seen as the \emph{weights} in artificial neural networks.
The variational quantum circuit approach has been shown to be flexible in circuit depth and somewhat resistant to noise~\cite{kandala2017hardware,farhi2014quantum,mcclean2016theory}.
Therefore,  even though there is still lack of quantum error correction and fault-tolerant quantum computation in the NISQ devices, the quantum machine learning algorithms powered by variational quantum circuits can circumvent the complex quantum errors which exist in the available quantum devices. Previous results in \cite{Farhi2018ClassificationProcessors, Mitarai2018QuantumLearning, Schuld2018Circuit-centricClassifiers}
have demonstrated that the variational quantum circuits can model any function approximators, classifiers and even quantum-many-body physics that are intractable on classical computers. For example, the work in \cite{Mitarai2018QuantumLearning} shows that a variational quantum circuits can approximate an analytical function $f(x)$.

It is hard to simulate quantum circuits of a large number of qubits via classical computers. 
\textcolor{black}{For example, a quantum circuit with $100$ qubits
  corresponds to a computational state space of dimensions  $2^{100}$.
 This huge number of computational state-space dimensions exceeds the storage and thus the computational capability of classical computers.
 Recently, Google demonstrated that a $53$-qubit quantum computer can successfully sample, by quantum measurement, one instance of the probability distribution of a quantum circuit a million times in around $200$ seconds while
it is estimated that such calculation to generate such amount of large size of random numbers will take $10000$ years on state-of-the-art classical supercomputer~\cite{arute2019quantum}.
In this study, we consider a small-scale simulation to demonstrate the possibility of running DRL applications on a quantum computer. The scale we consider here (several qubits) is still simulable by a quantum simulator on a classical computer. The hope is that when a larger scale, for example, $100$-qubit quantum computer is available, we may be, with some small changes to the variational quantum circuits, able to implement a VQ-DQL or VQ-DQN agent that is impossible to be simulated on a classical computer. For a review on the advantages of quantum computers over classical computers, see~\cite{harrow2017quantum}.  } 
In this work, we attempt to expand the expressive power of variational quantum circuits for the action-value function of DRL. \textcolor{black}{In certain cases, the variational quantum circuits require fewer parameters than a conventional neural network~\cite{du2018expressive}, making them promising for modeling complex environments. Consider a physical system with a size of \textcolor{black}{$100$ qubits;} it is basically impossible to simulate this system on any currently available supercomputer, or this system requires a significant amount of classical computing resources beyong what is currently available to simulate.
 Therefore, the expressive power we consider here is that potentially some applications of a very large size may be represented either by a quantum circuit of an intermediate size of qubits or in theory by a classical neural network while the quantum circuit would require a fewer number of parameters than the classical neural network.   }

\begin{figure}[H]
\begin{center}
\begin{minipage}{10cm}
\Qcircuit @C=1em @R=1em {
\lstick{\ket{0}} & \gate{R_x(\theta_1)} & \gate{R_z(\phi_1)} & \ctrl{1}   & \qw       & \qw      & \gate{R(\alpha_1, \beta_1, \gamma_1)} & \meter \qw \\
\lstick{\ket{0}} & \gate{R_x(\theta_2)} & \gate{R_z(\phi_2)} & \targ      & \ctrl{1}  & \qw      & \gate{R(\alpha_2, \beta_2, \gamma_2)} & \meter \qw \\
\lstick{\ket{0}} & \gate{R_x(\theta_3)} & \gate{R_z(\phi_3)} & \qw        & \targ     & \ctrl{1} & \gate{R(\alpha_3, \beta_3, \gamma_3)} & \meter \qw \\
\lstick{\ket{0}} & \gate{R_x(\theta_4)} & \gate{R_z(\phi_4)} & \qw        & \qw       & \targ    & \gate{R(\alpha_4, \beta_4, \gamma_4)} & \meter \gategroup{1}{4}{4}{7}{.7em}{--}\qw 
}
\end{minipage}
\end{center}
\caption[Generic circuit architecture for the variational quantum deep reinforcement learning.]{{\bfseries Generic variational quantum circuit architecture for the deep $Q$ network (VQ-DQN).}
  The single-qubit gates $R_x(\theta)$ and $R_z(\theta)$ represent rotations along $x$-axis and $z$-axis by the given angle $\theta$, respectively. The CNOT gates are used to entangle quantum states from each qubit and $R(\alpha,\beta,\gamma)$ represents the general single qubit unitary gate with three parameters.
  The parameters labeled $\theta_i$ and $\phi_i$ are for state preparation and are not subject to iterative optimization.  Parameters labeled $\alpha_i$, $\beta_i$ and $\gamma_i$ are the ones for iterative optimization.  Note that the number of qubits can be adjusted to fit the problem of interest and the grouped box may repeat several times to increase the number of parameters, subject to the capacity and capability of the quantum machines used for the experiments.}
\label{Fig:VQC_DQN}
\end{figure}
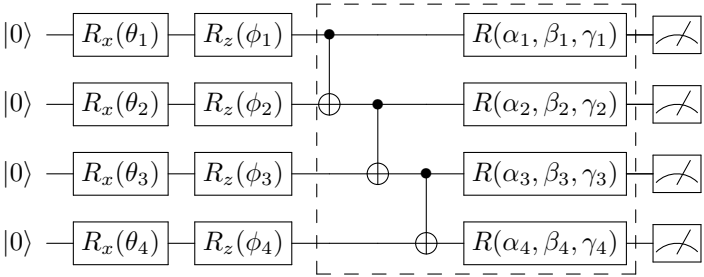

\begin{figure}[htbp]
     \centering
     \begin{subfigure}[b]{0.4\textwidth}
     \scalebox{0.85}{
     \begin{minipage}[b]{.3\textwidth}
            \Qcircuit @C=1em @R=1em {
            \lstick{\ket{0}} & \gate{R_x(\theta_1)} & \gate{R_z(\phi_1)} & \ctrl{1}   & \qw       & \qw      & \gate{R(\alpha_1, \beta_1, \gamma_1)} & \meter \qw \\
            \lstick{\ket{0}} & \gate{R_x(\theta_2)} & \gate{R_z(\phi_2)} & \targ      & \qw  & \qw      & \gate{R(\alpha_2, \beta_2, \gamma_2)} & \meter \gategroup{1}{4}{2}{7}{.7em}{--}\qw
            }
        \end{minipage}
        }
        
         \caption{Two Channels}
         \label{fig:VQDQN_Two_Channels}
     \end{subfigure}
     \hfill
     \begin{subfigure}[b]{0.4\textwidth}
     \scalebox{0.8}{
     \begin{minipage}[b]{.3\textwidth}
            \Qcircuit @C=1em @R=1em {
                \lstick{\ket{0}} & \gate{R_x(\theta_1)} & \gate{R_z(\phi_1)} & \ctrl{1}   & \qw       & \qw      & \gate{R(\alpha_1, \beta_1, \gamma_1)} & \meter \qw \\
                \lstick{\ket{0}} & \gate{R_x(\theta_2)} & \gate{R_z(\phi_2)} & \targ      & \ctrl{1}  & \qw      & \gate{R(\alpha_2, \beta_2, \gamma_2)} & \meter \qw \\
                \lstick{\ket{0}} & \gate{R_x(\theta_3)} & \gate{R_z(\phi_3)} & \qw        & \targ     & \ctrl{1} & \gate{R(\alpha_3, \beta_3, \gamma_3)} & \meter \qw \\
                \lstick{\ket{0}} & \gate{R_x(\theta_4)} & \gate{R_z(\phi_4)} & \qw        & \qw       & \targ    & \qw & \qw \gategroup{1}{4}{4}{7}{.7em}{--}\qw 
                }
            \end{minipage}
     }
        
         \caption{Three Channels}
         \label{fig:VQDQN_Three_Channels}
     \end{subfigure}
     
     \begin{subfigure}[b]{0.4\textwidth}
     \scalebox{0.8}{
     \begin{minipage}[b]{.3\textwidth}
            \Qcircuit @C=1em @R=1em {
                \lstick{\ket{0}} & \gate{R_x(\theta_1)} & \gate{R_z(\phi_1)} & \ctrl{1}   & \qw       & \qw      & \gate{R(\alpha_1, \beta_1, \gamma_1)} & \meter \qw \\
                \lstick{\ket{0}} & \gate{R_x(\theta_2)} & \gate{R_z(\phi_2)} & \targ      & \ctrl{1}  & \qw      & \gate{R(\alpha_2, \beta_2, \gamma_2)} & \meter \qw \\
                \lstick{\ket{0}} & \gate{R_x(\theta_3)} & \gate{R_z(\phi_3)} & \qw        & \targ     & \ctrl{1} & \gate{R(\alpha_3, \beta_3, \gamma_3)} & \meter \qw \\
                \lstick{\ket{0}} & \gate{R_x(\theta_4)} & \gate{R_z(\phi_4)} & \qw        & \qw       & \targ    & \gate{R(\alpha_4, \beta_4, \gamma_4)} & \meter \gategroup{1}{4}{4}{7}{.7em}{--}\qw 
                }
            \end{minipage}
            }
        
         \caption{Four Channels}
         \label{fig:VQDQN_Four_Channels}
     \end{subfigure}
     \hfill
     \begin{subfigure}[b]{0.4\textwidth}
        \scalebox{0.75}{
        \begin{minipage}[b]{.3\textwidth}
            \Qcircuit @C=1em @R=1em {
            \lstick{\ket{0}} & \gate{R_x(\theta_1)} & \gate{R_z(\phi_1)} & \ctrl{1}   & \qw       & \qw      & \qw & \gate{R(\alpha_1, \beta_1, \gamma_1)} & \meter \qw \\
            \lstick{\ket{0}} & \gate{R_x(\theta_2)} & \gate{R_z(\phi_2)} & \targ      & \ctrl{1}  & \qw      & \qw & \gate{R(\alpha_2, \beta_2, \gamma_2)} & \meter \qw \\
            \lstick{\ket{0}} & \gate{R_x(\theta_3)} & \gate{R_z(\phi_3)} & \qw        & \targ     & \ctrl{1} & \qw & \gate{R(\alpha_3, \beta_3, \gamma_3)} & \meter \qw \\
            \lstick{\ket{0}} & \gate{R_x(\theta_4)} & \gate{R_z(\phi_4)} & \qw        & \qw       & \targ    & \ctrl{1} & \gate{R(\alpha_4, \beta_4, \gamma_4)} & \meter \qw \\
            \lstick{\ket{0}} & \gate{R_x(\theta_5)} & \gate{R_z(\phi_5)} & \qw        & \qw       & \qw      & \targ & \gate{R(\alpha_5, \beta_5, \gamma_5)} & \meter \gategroup{1}{4}{5}{8}{.7em}{--}\qw 
            }
            \end{minipage}
            }
        
         \caption{Five Channels}
         \label{fig:VQDQN_Five_Channels}
     \end{subfigure}
        \caption{{\bfseries Variational quantum circuits for the cognitive-radio experiments.} The basic architecture is the same as the circuit used in the frozen-lake experiment. The main difference is that we have different number of qubits in order to fit the number of channels. In cognitive-radio experiments, all circuits are with two layers (grouped box repeated twice), regardless of the number of channels.\textcolor{black}{In the $3$-channel case, since the number of possible state is $3^2$, which is greater than the number of computational basis of a $3$-qubit system (which is $2^3 = 8$). We use a $4$-qubit system to accommodate the all possible states while the number of parameters is not increased.} }
        \label{fig:VQC_For_Cognitive_Radio}
\end{figure}
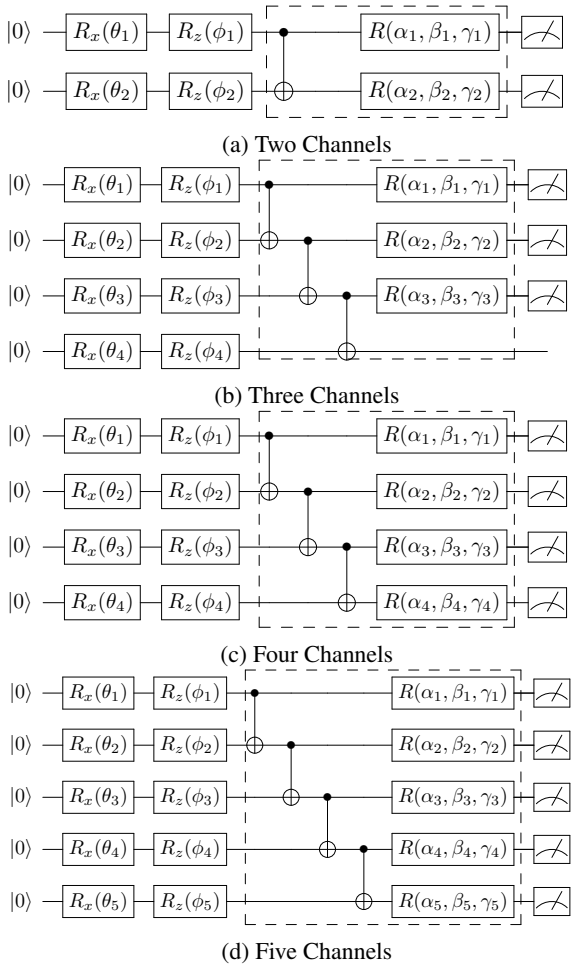

In~\figureautorefname{\ref{VQDQN_Overview}}, we present the overview of our proposed variational quantum circuit based DRL and its relevant components.
The RL agent includes a quantum part and a classical part.
The quantum part 
of the variational quantum circuit takes two sets of parameters and outputs measurement results that determine possible actions to take.
The classical part of a classical computer performs
the optimization procedure and calculates what new sets of parameters should be.
\figureautorefname{\ref{Fig:VQC_DQN}} shows a generic quantum circuit architecture for DRL (the detailed description of the quantum circuit will be presented later), and the algorithm for the VQ-DQL or VQ-QDN is presented in Algorithm 1. 
We construct two sets of circuit parameters with the same circuit architecture. The main circuit parameters are updated every step, while the target circuit parameters are updated per $20$ steps. For \emph{experience replay}, the replay memory is set for the length of $80$ to adapt to the frozen-lake testing environment and the length of $1000$ for the
cognitive-radio testing environment, and the size of training batch is $5$ for all of the environments.
The process of optimization needs to calculate gradients of expectation values of quantum measurements, which can be conducted by the same circuit architecture and slightly different parameters, respectively~\cite{Schuld2019EvaluatingHardware}.
Further, we encode the state with \emph{computational basis encoding}. 
In the frozen-lake
environment~\cite{Brockman2016OpenAIGym} we consider, there are totally $16$ states. Thus, it requires $4$ qubits to represent all states (see~\figureautorefname{ \ref{Fig:VQC_DQN}}). In the cognitive-radio experiments, we apply similar method and circuit architectures with different number of qubits to match the number of possible channels (see~\figureautorefname{ \ref{fig:VQC_For_Cognitive_Radio}}). Besides, ~\cite{Schuld2018InformationEncoding} provides a general discussion about the different encoding schemes. We discuss next the concept of computational basis encoding and the quantum circuits for the frozen-lake and cognitive-radio problems.

\begin{algorithm*}[t]
\begin{algorithmic}
\State Initialize replay memory $\mathcal{D}$ to capacity $N$
\State Initialize action-value function quantum circuit $Q$ with random parameters
\For{episode $=1,2,\ldots,M$} 
\State Initialise state $s_1$ and encode into the quantum state
\For {$t=1,2,\ldots,T$}
	\State With probability $\epsilon$ select a random action $a_t$
	\State otherwise select $a_t = \max_{a} Q^*(s_t, a; \theta)$ from the output of the quantum circuit
	\State Execute action $a_t$ in emulator and observe reward $r_t$ and next state $s_{t+1}$
	\State Store transition $\left(s_t,a_t,r_t,s_{t+1}\right)$ in $\mathcal{D}$
	\State Sample random minibatch of transitions $\left(s_j,a_j,r_j,s_{j+1}\right)$ from $\mathcal{D}$
	\State Set
	$y_j =
    \left\{
    \begin{array}{l l}
      r_j  \quad & \text{for terminal } s_{j+1}\\
      r_j + \gamma \max_{a'} Q(s_{j+1}, a'; \theta) \quad & \text{for non-terminal } s_{j+1}
    \end{array} \right.$
	\State Perform a gradient descent step on $\left(y_j - Q(s_j, a_j; \theta) \right)^2$
\EndFor
\EndFor
\end{algorithmic}
\caption{Variational Quantum Deep Q Learning}
\label{alg}
\end{algorithm*}

\noindent
\textcolor{black}{It has been shown that artificial neural networks~(ANN) are \emph{universal approximators}~\cite{hornik1989multilayer}, meaning that in theory, a single hidden layer neural network can approximate any computable function. However, the number of neurons in this hidden layer may be very large, which means that this model contains so many parameters. In machine learning applications, in addition to the capability of approximation, one needs to consider the amount of resources the model consumes. }

\subsection{Computational Basis Encoding and  Quantum Circuit for the Frozen Lake Problem}\label{sec:FL-VQC}
A general $n$-qubit state can be represented as:
\begin{equation}
    \ket{\psi} = \sum_{(q_1,q_2,...,q_n) \in \{ 0,1\}^n}^{} c_{q_1,...,q_n}\ket{q_1} \otimes \ket{q_2} \otimes \ket{q_3} \otimes ... \otimes \ket{q_n},
\end{equation}
where $ c_{q_1,...,q_n} \in \mathbb{C}$ is the \emph{amplitude} of each quantum state and each $q_n \in \{0,1\}$. 
The square of the amplitude $c_{q_1,...,q_n}$ is the \emph{probability} of measurement with the post-measurement state in  $\ket{q_1} \otimes \ket{q_2} \otimes \ket{q_3} \otimes ... \otimes \ket{q_n}$, and the total probability should sum to $1$, i.e.,
\begin{equation} \sum_{(q_1,q_2,...,q_n) \in \{ 0,1\}^n}^{} ||c_{q_1,...,q_n}||^2 = 1. 
\end{equation}

\textcolor{black}{We discuss the procedure, adopted here, to encode classical states  to the quantum register of the variational quantum circuit.
  Let us take the frozen-lake environment \cite{Brockman2016OpenAIGym} shown in~\figureautorefname{\ref{VQDQN_Env_FrozenLake}} as an example.
 There are $4\times 4=16$ possible states and    
 we label each possible state with an integer in the range from $0$ to $15$.
 We need a $4$-qubit system to include all possibilities of 16 states.
 The \emph{decimal} number is first converted into a \emph{binary} number and then encoded into the quantum states through single-qubit unitary rotations.
 In other words, each classical state can be denoted by a four-digit binary number $b_1b_2b_3b_4$, where $b_1, b_2, b_3, b_4$ can only take the value of $0$ or $1$, and then the corresponding encoded quantum state will be $\ket{b_1} \otimes \ket{b_2} \otimes \ket{b_3} \otimes \ket{b_4}$.
}

We propose the following single-qubit unitary rotation method to encode the classical input states from the testing environment into the quantum circuit of \figureautorefname{\ref{Fig:VQC_DQN}}.
In quantum computing, the single-qubit gate with rotation along the $j$-axis by angle $\theta$ is given by
\begin{equation}
  \label{eq:sqg}
  R_j(\theta)=e^{-i\theta\sigma_j/2}=\cos\frac{\theta}{2} I-i\sin\frac{\theta}{2}\sigma_j,
\end{equation}
where $I$ is an identity matrix and $\sigma_j$ is the Pauli matrix with $j=x,y,z$. 
The rotation angles for $R_x(\theta_i)$ and $R_z(\phi_i)$ in  \figureautorefname{ \ref{Fig:VQC_DQN}} are chosen to be
\begin{align} 
\theta_i &= \pi \times b_i , \label{theta_i}\\ 
\phi_i &= \pi \times b_i, \label{phi_i}
\end{align}
where $i$ represents the index of qubit $i$ and $\pi$ here is the radian. In the quantum circuit with four input qubits, the index is the set $\{1,2,3,4\}$.
The rotational angle parameters $\theta_i$ and $\phi_i$ are for state preparation and are not subject to iterative optimization.


\textcolor{black}{Take the state labeled $11$ observed by the agent as an example.
The decimal number $11$ of the state is first converted to the binary number $1011$, and then this classical state will be encoded into a quantum state $\ket{1} \otimes \ket{0} \otimes \ket{1} \otimes \ket{1}=|1011\rangle$.
The detailed procedure is as follows.
In this case, the binary digits  $b_1, b_2, b_3, b_4$ are $1,0,1,1$ respectively.
Then according to Eqs.~(\ref{theta_i}) and (\ref{phi_i}), 
One can obtain the values of $\theta_i$ and $\phi_i$
as $(\theta_1,\theta_2,\theta_3,\theta_4)=(\pi,0,\pi,\pi)$ and
$(\phi_1,\phi_2,\phi_3,\phi_4)=(\pi,0,\pi,\pi)$.
One can furthermore obtain from Eq.~(\ref{eq:sqg}) 
\begin{align} 
R_x(\pi) &= -i\sigma_x, \\
R_z(\pi) &= -i\sigma_z,
\end{align}
and
\begin{align} 
R_x(0) &= I, \\
R_z(0) &= I.
\end{align}
When the two quantum gates for encoding, $R_z(\theta_i) R_x(\phi_i)$,
operate on each qubit in initial state $\ket{0}$ as shown in \figureautorefname{ \ref{Fig:VQC_DQN}}, the resultant qubit state becomes  either
\begin{align} 
R_z(\pi)R_x(\pi)\ket{0}=(-i\sigma_z)(-i\sigma_x)\ket{0} &= \ket{1}
\end{align}
or
\begin{align} 
R_x(0)R_x(0)\ket{0}=II\ket{0} &= \ket{0}.
\end{align}
Thus one obtains for $b_1, b_2, b_3, b_4$ being  $1,0,1,1$, respectively, a quantum state $\ket{1} \otimes \ket{0} \otimes \ket{1} \otimes \ket{1}$. Other classical states can be encoded into their corresponding quantum states in the same way.
  This procedure is applicable for all the experiments in this work, including cognitive-radio experiments, regardless of the number of qubits.}

In the quantum circuit, the controlled-NOT (CNOT) gates are used to entangle quantum states from each qubit. 
\begin{equation}
  \label{eq:sqUg}
R(\alpha_i,\beta_i,\gamma_i)=R_z(\alpha_i)R_y(\beta_i)R_z(\gamma_i) 
\end{equation}
represents the general single qubit unitary gate with three parameters.
Parameters labeled $\alpha_i$, $\beta_i$ and $\gamma_i$ are the ones for iterative optimization.

The variation quantum circuit is flexible in circuit depth. A shallow circuit that well represents the solution space can still achieve approximately the goal of certain tasks although a more sophisticated and deeper circuit
may have better performance. But it remains a challenge to choose the right effective circuit that can parametrize and represent the solution space well for a general task while maintaining a low circuit
depth and a low number of parameters~\cite{sim2019expressibility}.
 It has been empirically demonstrated that 
 a strongly entangling low-depth circuit has the 
potential powers and advantages to efficiently represent the solution space for some specific problems~\cite{sim2019expressibility, havlivcek2019supervised, Schuld2018Circuit-centricClassifiers, kandala2017hardware}.
Thus we design the strongly entangling circuit by appending a layer (i.e., the grouped box in dashed lines in~\figureautorefname{\ref{Fig:VQC_DQN}}) comprised of two-quibit CNOT gates and parametrized general single-qubit unitary gates. 
Note that the layer or grouped box in dashed lines may repeat several times to increase the expressibility, entangling capability and also the number of parameters~\cite{sim2019expressibility}.
To accommodate both the use of NISQ machines and the performance of the variational quantum circuits, in this work, the grouped box repeats two times regardless of the testing environments.
The number of qubits can be adjusted to fit the problem of interest and the capacity of the simulators or quantum machines. For example, in the frozen-lake experiments and also the four-channel cognitive-radio experiments that will be discussed later, there are four input qubits and the grouped circuit repeats twice. Therefore the total number of circuit parameters subject to optimization is $4 \times 3 \times 2 = 24$. It is often to add a bias after the quantum measurement, the length of the bias vector is the same as the number of qubits. The bias vector is also subject to optimization. Therefore, the total number of parameters in this example is $24+4=28$ which is also listed in Table~\ref{tab:qadv_2}.

\begin{table*}[t]
\centering
\caption{Comparison of the number of parameters in classical Q-learning and variational quantum deep Q network (VQ-DQN).}
\label{tab:qadv_2}
\begin{tabular}{|l|l|l|l|l|l|}
\hline
  Environment & 2-Channels & 3-Channels & 4-Channels & 5-Channels & Frozen-Lake \\ \hline
Q-Learning &$2\times2\times2$&$3\times3\times3$&$4\times4\times4$&$5\times5\times5$&$4\times4\times4$\\ \hline
VQ-DQN     &$2\times(3\times2+1)$&$3\times(3\times2+1)$&$4\times(3\times2+1)$&$5\times(3\times2+1)$&$4\times(3\times2+1)$\\ \hline
\end{tabular}
\end{table*}

\subsection{Quantum Circuit for Cognitive Radio Networks}
\label{sec:CR-VQC}
In the experiments on the \emph{cognitive radio}~\cite{gawlowicz2018ns3}, the total number of channels $n$ that can be selected by the agent at each time-step is known in advance. Since the occupied channel changes from time to time, it is necessary to include not only the \emph{channel} but also the \emph{temporal} information into the observation. The \emph{observation} is in the following form: $(channel,time)$.
In our experimental setting, we consider that the channel-changing by the primary user follows a simple periodic pattern with $n$ time steps in a full cycle
(see~\figureautorefname{\ref{VQDQN_Env_CognitiveRadio_3_configs}} for the $n=4$ case). Therefore, the number of possible states is $n^2$. 
\textcolor{black}{Similar to the circuit architecture of the frozen-lake experiments 
  in \figureautorefname{ \ref{Fig:VQC_DQN}},
  the variational quantum circuits for cognitive-radio experiments  are shown in \figureautorefname{\ref{fig:VQC_For_Cognitive_Radio}}. We choose different qubit numbers to accommodate possible $n^2$ states of the  $n$-channel cognitive-radio environment. Normally, $n$ qubits are used for the encoding and action (channel) selection of an $n$-channel cognitive-radio environment.
  However, the $3$-channel case is special since the number of possible state is $3^2$, which is greater than the number of computational basis states of $3$-qubit system (which is $2^3 = 8$). We thus use a $4$-qubit system to accommodate the all possible states. The scheme to encode the classical $n^2$ states into their corresponding quantum states in the cognitive-radio environment is the same as that in the frozen-lake experiment introduced in \sectionautorefname{\ref{sec:FL-VQC}} except that some quantum states are not used when $2^{n_q} > n^2$, where $n_q$ is the qubit number and $n$ is the channel number. }

In addition, at each time step, the agent can select one of the channels from the set of all possible channels, which is of number $n$.
\textcolor{black}{
 This corresponds exactly to the $n$ possible actions that the agent can select.
The action selection in our VQ-DQN scheme is determined through the expectation values of $n$ qubits, which will be discussed in
\sectionautorefname{\ref{sec:Env-setup}}.
Since $n_q>n$ for channel number $n=3$, we use the repeated quantum measurements of only the first three qubits for the estimation of the expectation values for channel or action selection.
At the same time, we would like to keep the number of parameters to scale as $n\times (3 \times 2 + 1)$. So for the special case of $n=3$, there is no single-qubit unitary operation with optimization parameters acting on the fourth qubit in the grouped box in  \figureautorefname{\ref{fig:VQC_For_Cognitive_Radio}}(b).
We note here that since the number of possible qubit states $2^{n}$, which is greter than $n^2$ for $n>5$, grows exponentially, leaving our variational quantum circuit the possibilities to deal with more complex scenarios with more possible states than the simple periodical channel-changing pattern with $n^2$ possible states when the number of channels $n$ becomes large.
}

\section{Data Availability}
The data that support the findings of this study are available in the GitHub repository, \url{https://github.com/ycchen1989/Var-QuantumCircuits-DeepRL}

\section{Experiments and Results}

\begin{figure}[!htbp]
\begin{subfigure}{1.\linewidth}
\center
\includegraphics[width=1.0\linewidth]{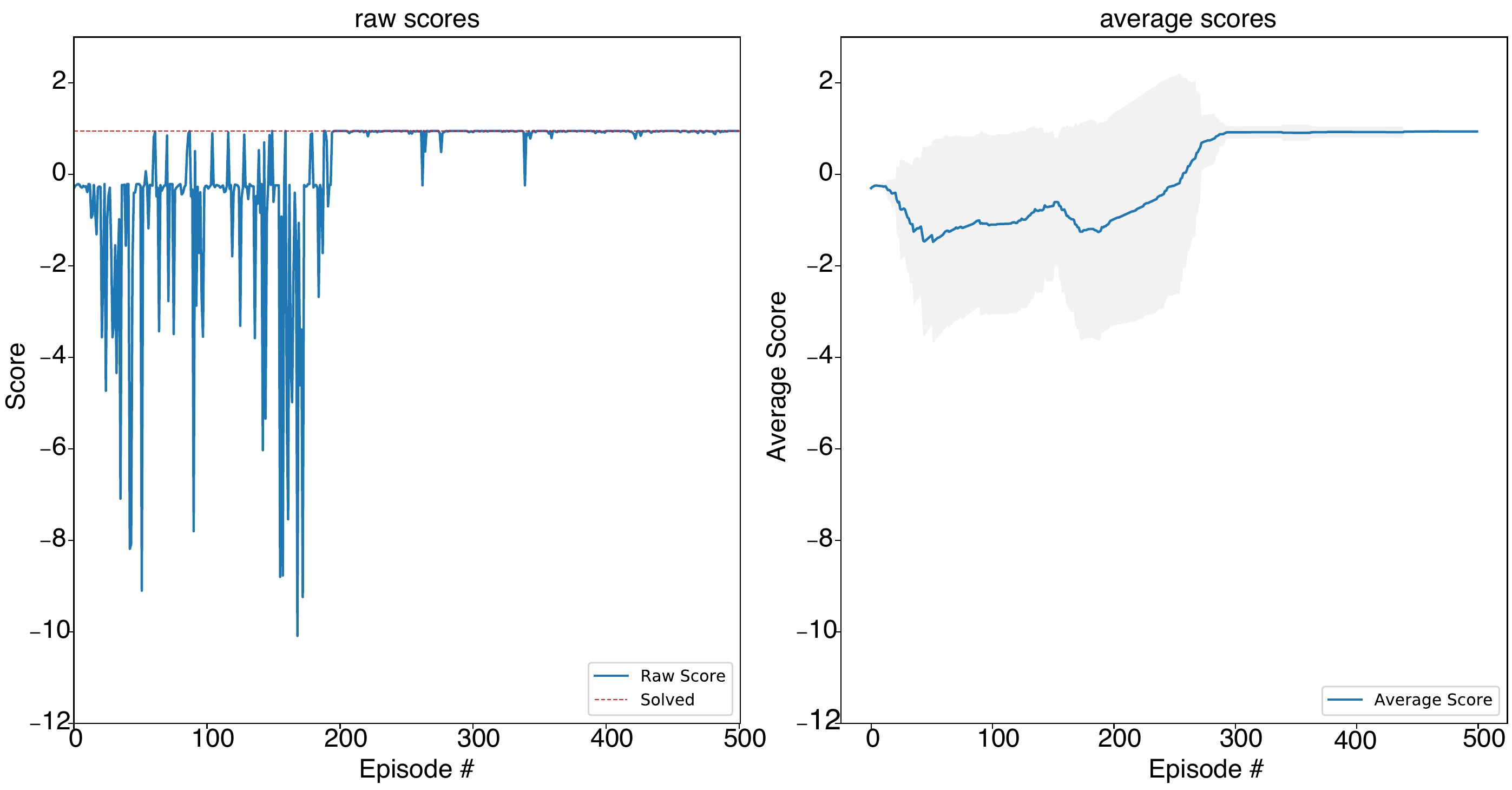}
\label{fig:frozenlake_original}
\subcaption{}
\end{subfigure}
\begin{subfigure}{1.\linewidth}
\center
\includegraphics[width=1.0\linewidth]{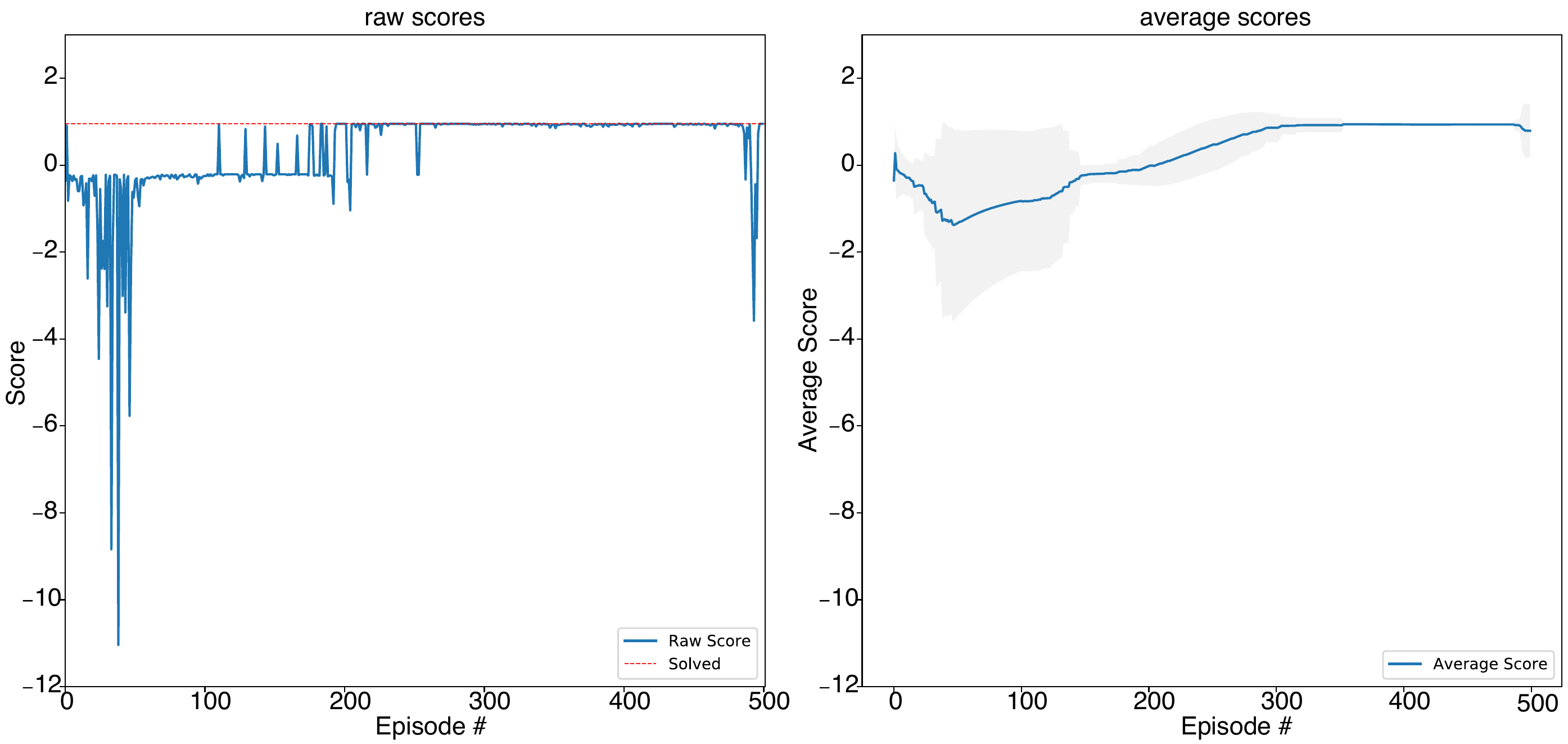}
\label{fig:frozenlake_config_1}
\subcaption{}
\end{subfigure}
\begin{subfigure}{1.\linewidth}
\center
\includegraphics[width=1.0\linewidth]{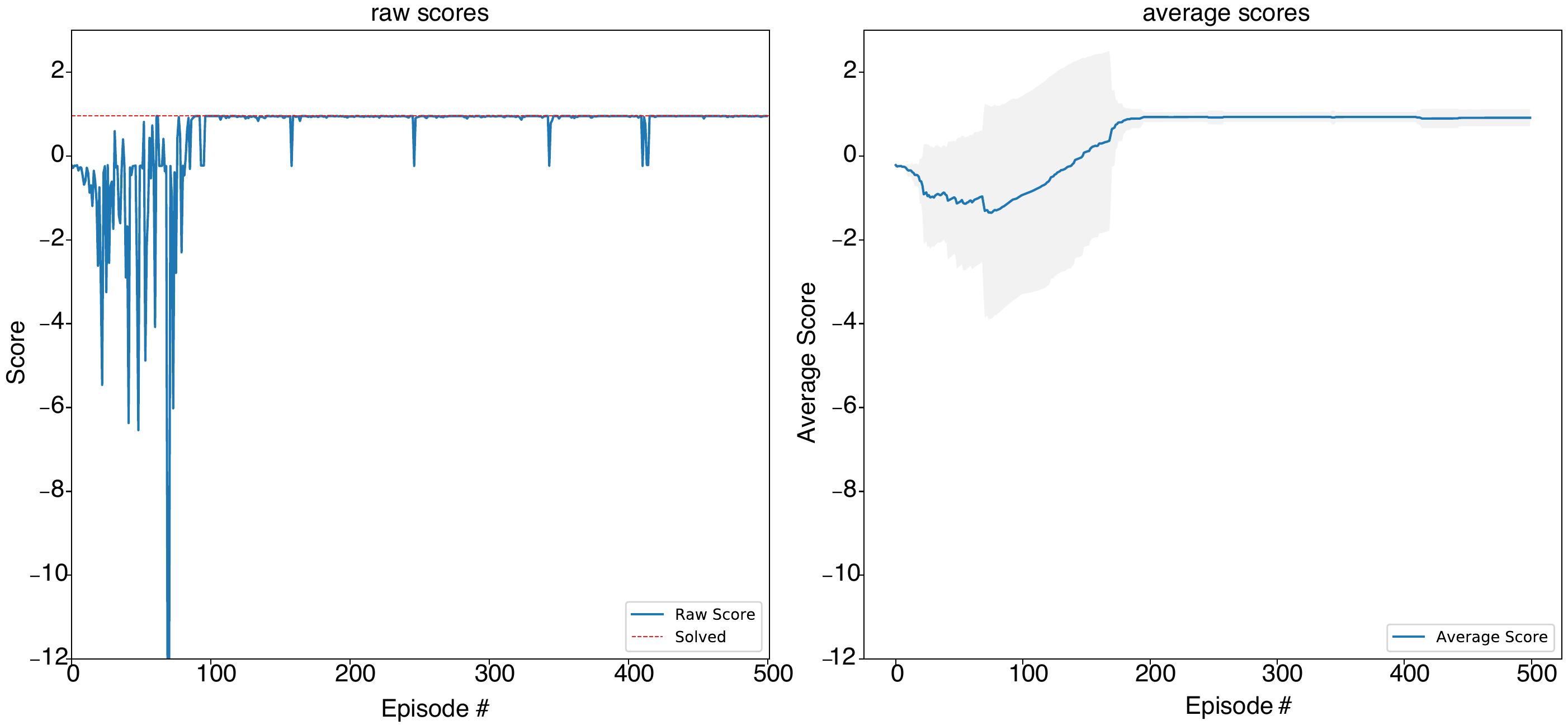}
\label{fig:frozenlake_config_4}
\subcaption{}
\end{subfigure}
\caption[Result:Variational quantum circuit for deep Q-learning (DQL)]{{\bfseries Performance of the variational quantum circuits for DQL on the frozen-lake experiment.} \textcolor{black}{Subfigures (a), (b), and (c) correspond to the results of the environment configurations (a), (b), and (c) in \figureautorefname{\ref{VQDQN_Env_FrozenLake}, respectively}.} \textcolor{black}{Take subfigure (a) for example. The left panel of subfigure (a)} shows that our proposed variational quantum circuits based DQL-agent reaches the optimal policy in the frozen-lake environment with a total reward of 0.9 at the 198th iteration. The gray area in the right panel of the subfigure represents the standard deviation of reward in each iteration during exploration with the standard reinforcement learning reproducible setting for stability. \textcolor{black}{The mean and the standard deviation values of the average score (reward) are calculated from the  scores (rewards) in the past $100$ episodes.} The policy learned via quantum circuit becomes more stable after the 301st iteration. \textcolor{black}{The other two experiments in subfigures (b) and (c) demonstrate the similar pattern.}}
\label{VQDQN_FrozenLake_CliffWalking}
\end{figure}


\begin{figure*}[htbp]
\center
\includegraphics[width=1.\linewidth]{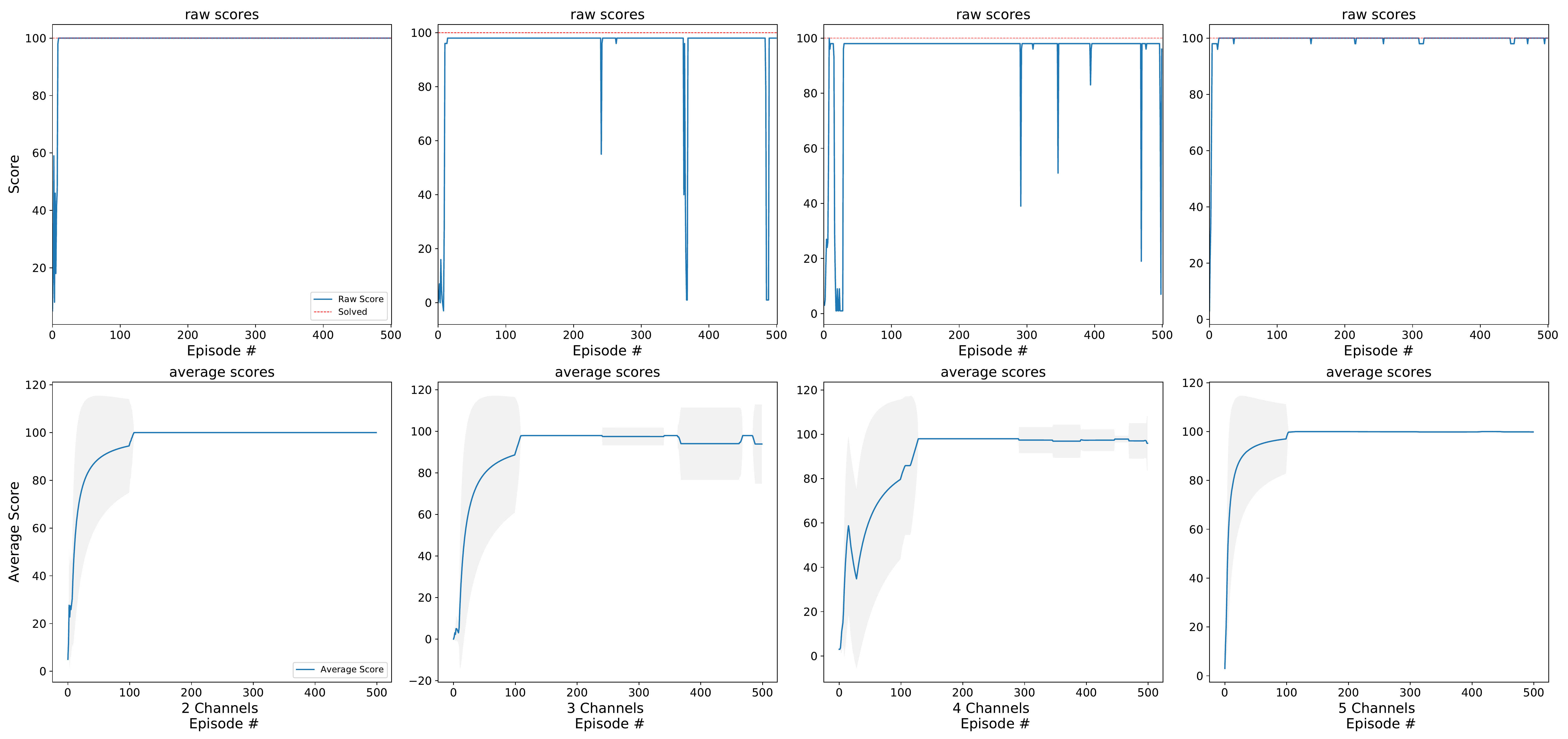}
\caption[Result:Variational Quantum Circuit for Deep Q-Learning (DQL)]{{\bfseries Performance of the Variational quantum circuits for DQL on the cognitive-radio experiment.}
In the cognitive-radio experiments, we limit the maximum steps an agent can run to be $100$, and the reward scheme is that for each correct choice of the channel, the agent will get a $+1$ reward and $-1$ for incorrect selection. The maximum reward an agent can achieve under this setting is $100$. The top panels of the figure show that our proposed variational quantum circuits based DQL-agent reaches the optimal policy with a total reward of $100$ in the 2-channel and 5-channel cases at only several iterations. For the cases of 3-channel and 4-channel, the agent also reaches near-optimal policy at only several iterations. The gray area in the bottom panels of the figure represents the standard deviation of reward in each iteration during exploration with the standard reinforcement learning reproducible setting for stability~\cite{9053342}. The mean and the standard deviation at each episode are calculated from the rewards(scores) in the past $100$ episodes. The policy learned via variational quantum circuits becomes more stable after around $100$ iterations for all the four cases.This part of experiment is tested on the configuration (a) shown in \figureautorefname{\ref{VQDQN_Env_CognitiveRadio_3_configs}}.}
\label{VQDQN_CognitiveRadio_2_Channels}
\end{figure*}

\begin{figure}[htbp]
\center
\includegraphics[width=1.\linewidth]{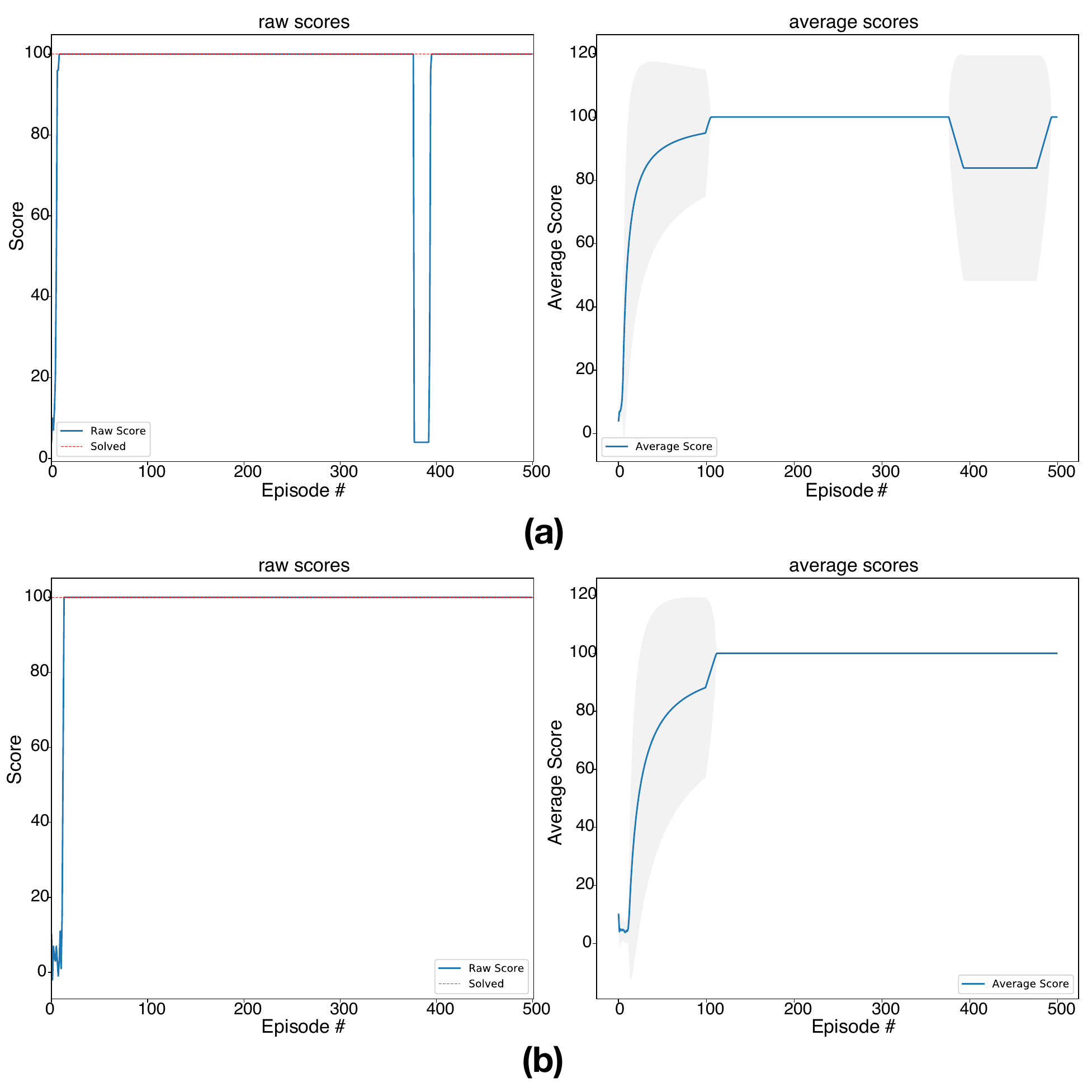}
\caption[Result:Variational Quantum Circuit for Deep Q-Learning (DQL)]{{\bfseries Performance of the Variational quantum circuits for DQL on the cognitive-radio experiment.}
\textcolor{black}{This part of experiment is tested on configurations (b) and (c) shown in \figureautorefname{\ref{VQDQN_Env_CognitiveRadio_3_configs}}, and the results are comparable to the results in~\figureautorefname{\ref{VQDQN_CognitiveRadio_2_Channels}}, which are tested on configuration (a) in \figureautorefname{\ref{VQDQN_Env_CognitiveRadio_3_configs}}. This demonstrates that our proposed quantum DRL (DQL) framework can be trained on different scenarios.} 
}
\label{VQDQN_CognitiveRadio_Additional_Configs}
\end{figure}

\begin{figure*}[htbp]
\center
\includegraphics[width=1.\linewidth]{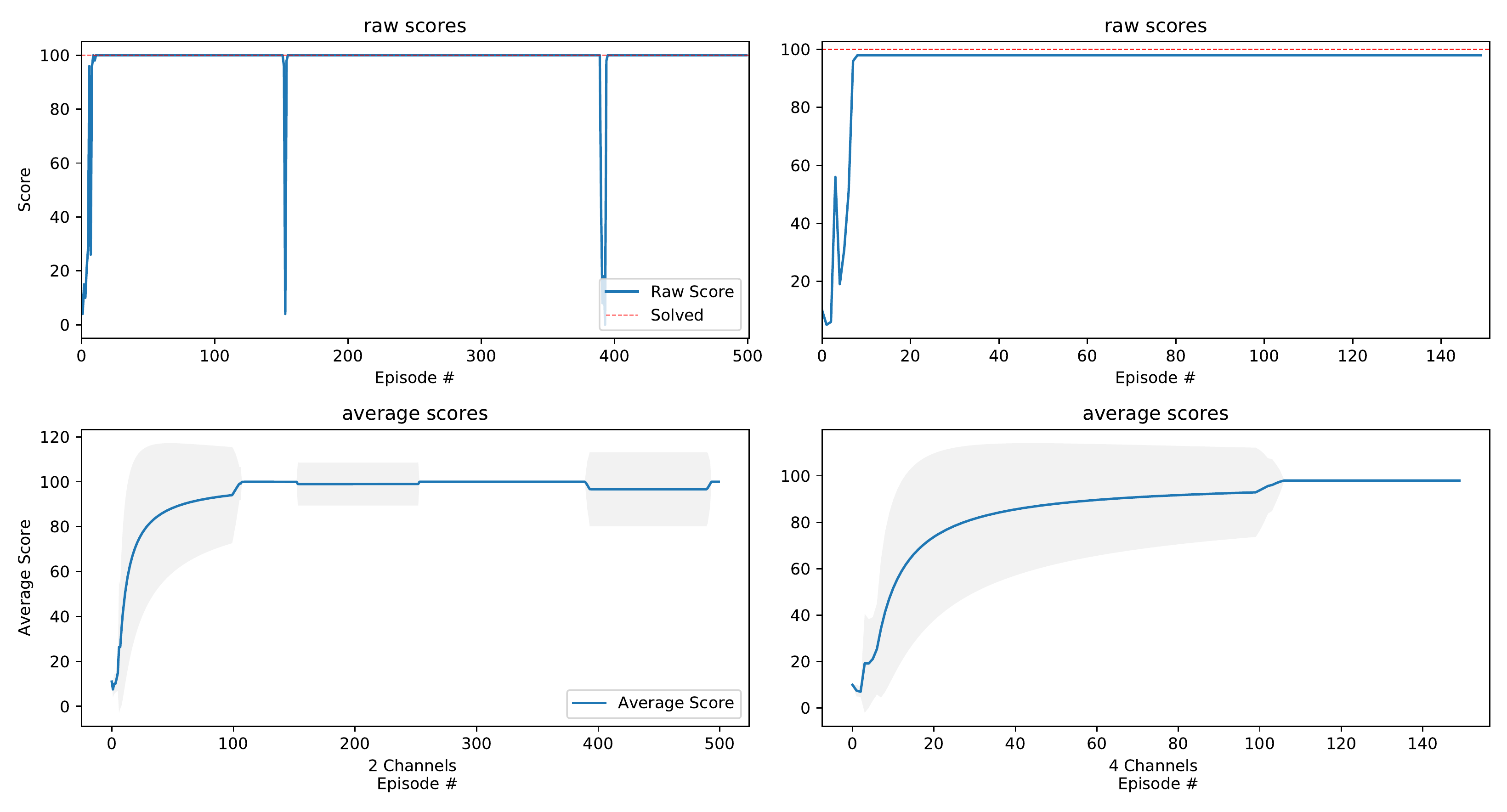}
\caption[Result:Variational Quantum Circuit for Deep Q Learning (DQL)]{{\bfseries Performance of the variational quantum circuit for DQL on cognitive-radio experiment conducted on the IBM Qiskit Aer quantum simulator with noise from the real quantum machine, \texttt{ibmq-poughkeepsie}.}
In this experiment, we investigated the robustness of the VQ-DQN against the noise in quantum machines. The device noise is downloaded from remote real quantum machines and then incorporated with the simulator software Qiskit Aer provided by IBM. This noisy backend then replaces the noiseless backend in previous experiments. The result of the experiment shows that our proposed variational quantum circuits based DQL or DQN is robust against noise in the current quantum machines. 
 }
\label{VQDQN_CognitiveRadio_2_3_4_5_With_IBM_NOISE}
\end{figure*}

\subsection{Environment Setup and Action Selection}
\label{sec:Env-setup}
The frozen-lake testing environment is depicted in
\figureautorefname{\ref{VQDQN_Env_FrozenLake}}.
We set up the experiment following the circuit architecture in \figureautorefname{ \ref{Fig:VQC_DQN}}.  
For the cognitive-radio environment, we consider
the case 
where the external interference of the primary user follows a periodic pattern, i.e. sweeping over all channels from $1$ to $n$ in the same order (see, e.g.,
\figureautorefname{\ref{VQDQN_Env_CognitiveRadio_3_configs}} for
the $n=4$ case).
We set up the experiment following the circuit architectures in \figureautorefname{  \ref{fig:VQC_For_Cognitive_Radio}} according to the number of channels $n$ in the cognitive-radio environment. 
This cognitive-radio environment offers a feasible test-bed for quantum DQL (DQN) with a desirable self-defined environment with lower action and space complexity working in the current NISQ machines.

Next, we describe how to select an action in the variational quantum circuit.
The measurement output of the expectation (ensemble average) values of the $n$-qubit quantum circuit is an $n$-tuple. The index counting of the  measurement qubit output ports or wires is from zero to $n-1$. The choice of the action is just the index of the measurement qubit output port or wire that has the largest expectation value.
Taking the four-qubit setting as an example, one then has the output of $[a ,b, c, d]$ with $a$, $b$, $c$, $d$ being numerical expectation values obtained from a set of measurements.
If the numerical value $b$ is the largest value among the four expectation values, then the action choice is $1$. This action $1$ will then be sent to the testing environment.
To be more specific, for the frozen-lake environment, there are four actions {LEFT, DOWN, RIGHT, UP} in the action space. 
The output ports or wires of the four-qubit quantum circuit are labeled $0,1,2,3$ and they correspond to the action {LEFT, DOWN, RIGHT, UP}, respectively. If output wire $1$ has the largest expectation value, then the action that will be selected by the agent is to go DOWN one step from the current state in the frozen-lake environment.  Similarly, there are $n$ possible action choices, Channel $1$ to Channel $n$, in the cognitive-radio environment, and they correspond to the output qubit ports or wires labeled from $0$ to $n-1$, respectively. If the output qubit wire $1$ has the largest measured expectation value, the agent (the secondary user) will select Channel $2$ as the action.

\textcolor{black}{
  As described above, the next action that the agent selects is determined by the expectation value of each qubit, not by the random outcome of $0$ or $1$ of each qubit in each single run of measurement. The expectation value can be calculated analytically if we use the quantum simulator (for example, PennyLane~\cite{Bergholm2018PennyLane:Computations}  or IBM Qiskit) on a classical computer, and in this case the result is deterministic. If the agent is on a real quantum computer, the expectation value is estimated with a large number of measurement samplings, which should be close to the value calculated theoretically by the quantum simulator.
Let us give a brief summary of the whole procedure on a real machine with an example. 
Suppose the RL agent receives a classical binary number for the state $1011$. First, this binary number will be converted into the quantum state $\ket{1} \otimes \ket{0} \otimes \ket{1} \otimes \ket{1}$ by the encoding single-qubit gates and then go through the quantum circuit blocks before the measurements. If we measure the first qubit, for example, we will get either $0$ or $1$, but which of these two possible outcomes will appear is truly stochastic. Thus this single-shot random measurement outcome
is not enough for the agent to decide the next move or action, and 
instead, the expectation value is used.
To find the expectation value, one needs to measure an ensemble of identically prepared systems, or perform an ensemble of measurements on a single quantum system repeatedly prepared in the identical quantum state $\ket{1} \otimes \ket{0} \otimes \ket{1} \otimes \ket{1}$, which is our case here.  
So the agent will 
prepare the qubits in state 
$\ket{1} \otimes \ket{0} \otimes \ket{1} \otimes \ket{1}$
and let it  go through the quantum circuit blocks and then perform a measurement on each qubit.
This process will repeated for a large number of times.
The measurement outcome of being in state $\ket{0}$ or state $\ket{1}$ 
in each measurement is purely random and unpredictable. However, the probability of being in state  $\ket{0}$ 
and  the probability of being in state $\ket{1}$  can be predicted.
The average value of all
these repeated measurements is the expectation value.
For example, if the agent repeats the process $1000$ times and get $600$ times of $1$ and $400$ times of $0$ on the first qubit, then the expectation value of the first qubit will be about $0.6$. The procedure is the same for other qubits when we consider multi-qubit systems.}

\textcolor{cyan}{
  Normally, the number of observations or measurements needed to learn a discrete probability distribution on an $n$-qubit system of
size $2^n$ is, in the worst case, linear in $2^n$. This means that learning the probabilities may require a number of measurements that scales exponentially with the number of qubits $n$. However, what is necessary for our algorithm to determine the next action by the QRL agent here is to find the expectation value of each qubit in the $n$-qubit system, rather than the expectation value on the $n$-qubit computational state space. As a result, we only need to measure discrete probability distribution of size $n$ rather than size $2^n$.
Moreover, the the measurements on each qubit of the
$n$ qubits can be performed simultaneously in parallel.
Thus the number of repeated experiments (measurements) to obtain the expectation values can be chosen to be a fixed number, rather independent of $n$.
Furthermore, our algorithm does not require to find the exact expectation value of each qubit, and only the qubit that has the largest expectation value is concerned, making our DRL algorithm relatively
robust against noise and errors in the real quantum machines. 
We note here that in the inference experiment with our trained QRL model running on a real quantum computer discussed in \sectionautorefname{\ref{sec:runningOnRealQuantumComputer}}, the number of repeated measurements is set to 1024, and this number of measurement shots can already give a fairly stable result. }

\subsection{Numerical Simulation}

The quantum circuits as constructed in  ~\figureautorefname{\ref{Fig:VQC_DQN}} and ~\figureautorefname{\ref{fig:VQC_For_Cognitive_Radio}} are numerically simulated with the software package PennyLane \cite{Bergholm2018PennyLane:Computations}. We use the standard package PyTorch \cite{Paszke2017AutomaticPyTorch} to help the linear algebraic manipulation to accelerate the simulation. OpenAI Gym~\cite{Brockman2016OpenAIGym} provides the testing environment. In this work, we choose an environment with low computational complexity, the frozen-lake environment, to implement the proof-of-concept experiments and choose the cognitive-radio environment to study the possible real-world applications.

The optimization method is chosen to be \emph{RMSprop}~\cite{Tieleman2012} in which parameters are set to be $\emph{learning rate}=0.01$, $\emph{alpha} = 0.99$ and $\emph{eps} = 10^{-8}$, widely used in DRL. Note that the learning rate, alpha and eps here are only for the gradient descent optimization. Please do not be confused with the aforementioned DRL hyperparameters. The batch-size for the experience replay is $5$. The $\epsilon$-greedy strategy used in the frozen-lake environment
 is the following:
\begin{equation}
\epsilon \xleftarrow{} \frac{\epsilon}{\frac{episode}{100} + 1},
\end{equation}
and in the cognitive-radio environment 
$\epsilon$ is updated in every single step as:
\begin{equation}
\epsilon \xleftarrow{} 0.99\epsilon
\end{equation}
with initial $\epsilon = 1.0$ for encouraging more exploration in early episodes and shifting to more exploitation in later episodes.



\subsection{Simulation with Noise}
\label{sec:simulationWithNoiseFromIBM}
To investigate the robustness of our proposed variational quantum circuit-based DRL against the noise from current and possible near-term NISQ devices, we perform additional simulations which include the noise from a real quantum computer. The experiment setting is the same as the previous experiments, except that the simulation backend is replaced with the Qiskit-Aer simulation software, which has the capability to incorporate the noise models from the IBM quantum computers. We perform the noise-model simulations for the 2-channel and 4-channel settings in cognitive-radio experiments. 
\textcolor{black}{The qubit properties, such as relaxation time $T_1$, dephasing time $T_2$, qubit frequency, gate error and gate length, of the 
  IBM Q $20$-qubit machine \texttt{ibmq-poughkeepsie}, from which we download noise data for the above mentioned noise-model simulations, are listed in Tables~\ref{tab:ibm_noise_simulation}, \ref{tab:ibm_gate_error_simulation},
  and \ref{tab:ibm_gate_coupling_error_simulation} in Appendix \ref{Noise:ibmq-poughkeepsie}.}

The variational quantum circuits can be relatively robust against noises because they involve a classical optimization step and the related deviations can be absorbed by the parameters during the iterative optimization process. 
\textcolor{cyan}{
  Another appealing feature of our quantum variational DRL algorithm is the additional ability to tolerate errors and noises. 
The next action that the agent selects in our algorithm 
is determined by which qubit has the largest expectation value among all the $n$ qubits. In other words, although quantum gate opertions and the measurement fidelity on NISQ devices are degraded by noises and errors,
as long as the qubit that has the largest expectation value is the same as that on the ideal quantum simulator, the DRL agent will select the same action.}

\subsection{Performance Analysis}
\label{sec:numericalSimulations}
\textcolor{black}{In the frozen-lake experiment, we run 500 episodes on all three configurations.
Subfigures (a), (b), and (c) in \figureautorefname{ \ref{VQDQN_FrozenLake_CliffWalking}} correspond to the results of the environment configurations (a), (b), and (c) in \figureautorefname{\ref{VQDQN_Env_FrozenLake}}, respectively.
  Take subfigure (a) for example. The agent converges to the total reward 0.9 after the 198th episode. The results of the other two configurations shown in subfigures (b) and (c) in \figureautorefname{ \ref{VQDQN_FrozenLake_CliffWalking}} also demonstrate the similar pattern. It is noted that, however, several sub-optimal results occurr. This phenomenon is probably due to the $\epsilon$-greedy policy selection.}
 
\textcolor{black}{
  To demonstrate the stability of our quantum RL agents' training process, we calculate the temporary mean value and the standard-deviation boundary of the total rewards. The mean and standard-deviation values are calculated from the last 100 episodes. In the case that there are fewer than 100 episodes, we include all the episodes. The standard deviation values are shown in gray color and the mean values are in blue color in the right panel of~\figureautorefname{\ref{VQDQN_FrozenLake_CliffWalking}}. In our simulations, the quantum RL agents' \textcolor{black}{average total reward (average score)} are with small standard-deviation values, meaning that they are stable after training.
}

\textcolor{black}{In the cognitive-radio experiment, we tested situations where there are $2$, $3$, $4$ and $5$ possible channels with the environment configuration shown in (a) of \figureautorefname{\ref{VQDQN_Env_CognitiveRadio_3_configs}} for the agent to choose, respectively.} In all the four situations, we run 500 episodes and the agent converges to the optimal total reward value at around 100 iterations or episodes (see \figureautorefname{ \ref{VQDQN_CognitiveRadio_2_Channels}}). 
\textcolor{black}{We further perform the simulations on the other two 4-channel environment configurations shown in (b) and (c) of \figureautorefname{\ref{VQDQN_Env_CognitiveRadio_3_configs}}, the training result shown in \figureautorefname{\ref{VQDQN_CognitiveRadio_Additional_Configs}} is comparable to the 4-channel training result on environment configuration (a) shown in \figureautorefname{ \ref{VQDQN_CognitiveRadio_2_Channels}}.}
\textcolor{black}{In the simulation with the noise model from the IBM quantum machine (with the environment configuration shown in (a) of \figureautorefname{\ref{VQDQN_Env_CognitiveRadio_3_configs}}),} the agent converges to the optimal total reward value at around 110 iterations in the 2-channel and 4-channel settings (see \figureautorefname{\ref{VQDQN_CognitiveRadio_2_3_4_5_With_IBM_NOISE}}), which are comparable to previous ideal and noiseless simulations in  \figureautorefname{ \ref{VQDQN_CognitiveRadio_2_Channels}}, showing that our proposed quantum circuit based DRL is robust against noise in the current machines.
\textcolor{black}{
  The mean and standard-deviation values of the total rewards for cognitive-radio experiments are shown 
  in the right panel of \figureautorefname{\ref{VQDQN_CognitiveRadio_Additional_Configs}} and the bottom panels of \figureautorefname{\ref{VQDQN_CognitiveRadio_2_Channels}} and \figureautorefname{\ref{VQDQN_CognitiveRadio_2_3_4_5_With_IBM_NOISE}}. The small standard-deviation values indicate that they are stable after training. 
}


\subsection{Running On a Quantum Computer}
\label{sec:runningOnRealQuantumComputer}

\begin{table}[h]
\centering
\caption{{\bfseries  \textcolor{black}{Results of the trained VQ-DQL (VQ-DQN) for the cognitive-radio experiment conducted on the IBM Q quantum computer, \texttt{ibmq-valencia}.}}
  \textcolor{black}{
    In this experiment, we test the trained quantum DRL model in the cognitive-radio experiment of configuration (c) in the $4$-channel case described in~\figureautorefname{\ref{VQDQN_Env_CognitiveRadio_3_configs}}. Even if the training is on the simulation software without the quantum noise, the trained model still performs well on the real quantum computer.} }
\begin{tabular}{|c|c|c|c|c|c|}
\hline
Episodes     & 1   & 2   & 3   & 4 & 5\\ \hline
Total Steps  & 100 & 100 & 100 & 100 & 100 \\ \hline
Total Reward & 100 & 100 & 100 & 100 & 98 \\ \hline
\end{tabular}
\label{tab:ns3_4_channels_IBMQ}
\end{table}

\begin{table}[h]
\centering
\caption{{\bfseries  \textcolor{black}{Results of the trained VQ-DQL (VQ-DQN) for the frozen-lake experiment conducted on the IBM Q quantum computer, \texttt{ibmq-valencia}.}}
  \textcolor{black}{
    In this experiment, we test the trained quantum DRL model in the frozen-lake experiment of configuration (c) in~\figureautorefname{\ref{VQDQN_Env_FrozenLake}}. Even if the training is on the simulation software without the quantum noise, the trained model still performs well on the real quantum computer.}}
\begin{tabular}{|c|c|c|c|c|c|c|c|}
\hline
Episode     & 1   & 2   & 3   & 4 & 5 & 6 & 7\\ \hline
Total Steps  & 6 & 6 & 6 & 7 & 7 & 7  & 6 \\ \hline
Total Reward & 0.95 & 0.95 & 0.95 & 0.94 & 0.94 & 0.94 & 0.95\\ \hline
\end{tabular}
\label{tab:frozen_lake_config_4_IBMQ}
\end{table}

\textcolor{black}{We further upload our trained VQ-DQL (VQ-DQN) models to the IBM Q cloud-based quantum computing platform to investigate whether the models are feasible on a real quantum computer. In the cognitive-radio experiment, we upload the trained model parameters of the variational quantum circuit of configuration (c) in the $4$-channel case described in~\figureautorefname{\ref{VQDQN_Env_CognitiveRadio_3_configs}}.
  Due to the limited resource available on the cloud-based quantum computing platform, we exclusively carry out five episodes of this specific experiment on the IBM Q backend machine  \texttt{ibmq-valencia}. The result of the experiment conducted on the IBM Q machine 
listed in~\tableautorefname{\ref{tab:ns3_4_channels_IBMQ}}
has almost the same total reward as that obtained by running on the PennyLane or Qiskit quantum simulator. This demonstrates that even if the training is on the simulation software without the noise, the trained model of the variational quantum circuit for DRL still performs well on the real quantum computer.
\textcolor{cyan}{The reason for this is that our quantum DRL algarithm does not require to find the exact expectation value of each qubit, and only cares which qubit that has the largest expectation value. Thus, our algorithm is relatively robust against errors and noises.}
In the frozen-lake experiment, we test the trained model parameters for the environment configuration (c) described in \figureautorefname{\ref{VQDQN_Env_FrozenLake}} also on the IBM Q machine \texttt{ibmq-valencia} and carry out seven episodes of this specific experiment. The result listed in~\tableautorefname{\ref{tab:frozen_lake_config_4_IBMQ}} for the quantum DRL model experiment conducted on the IBM Q quantum computer is also comparable to that obtained by running on the PennyLane or Qiskit quantum simulator.  In these real-machine experiments, the number of shots of quantum measurements for the calculation or estimation of the expectation values is $1024$.
The qubit properties, such as relaxation time $T_1$, dephasing time $T_2$ and qubit frequency, gate error and gate length, of the 
IBM Q $5$-qubit machine  \texttt{ibmq-valencia} we use for the above mentioned two experiments are listed in Tables~\ref{tab:ibm_noise_valencia}, \ref{tab:ibm_gate_error_valencia}, 
and \ref{tab:ibm_coupling_gate_error_valencia} in Appendix \ref{Noise:ibmq-valencia}. }

\subsection{Quantum Advantage on Memory Consumption}

\begin{table*}[t]
\centering
\caption{Comparison of classical reinforcement learning algorithms with discrete action space and variational quantum deep Q networks (VQ-DQN)
}
\label{tab:qadv}
\begin{threeparttable}
\begin{tabular}{|c|ccccc|}
\hline
\textbf{Algorithm} & \multicolumn{1}{c|}{\textbf{Policy}} & \multicolumn{1}{c|}{\textbf{Action Space}} & \multicolumn{1}{c|}{\textbf{State Space}} & \multicolumn{1}{c|}{\textbf{Operator}} & \textbf{Complexity of Parameters} \\ \hline
Monte Carlo & Off-policy & Discrete & Discrete & Sample-means &  $\mathcal{O}(n^3)$\\ \hline
Q-Learning & Off-policy & Discrete & Discrete & Q-value &  $\mathcal{O}(n^3)$\\ \hline
SARSA & On-policy & Discrete & Discrete & Q-value &  $\mathcal{O}(n^3)$\\ \hline
DQN & Off-policy & Discrete & Continuous & Q-value & $\mathcal{O}(n^2)$ \\ \hline
VQ-DQN \tnote{a} & Off-policy & Quantum & Quantum & Q-value &  $\mathcal{O}(n)$\\ \hline
VQ-DQN \tnote{b} & Off-policy & Quantum & Quantum & Q-value &  $\mathcal{O}(\log{} n)$\\ \hline
\end{tabular}
\begin{tablenotes}
            \item[a] The number of parameters in VQ-DQN with computational basis encoding grows only linearly with the dimension of the input vector $n$.
            \item[b] VQ-DQN with amplitude encoding can harvest full logarithmic less parameters compared with classical models.
        \end{tablenotes}
\end{threeparttable}
\end{table*}


\begin{figure}[H]
\center
\includegraphics[width=0.9\linewidth]{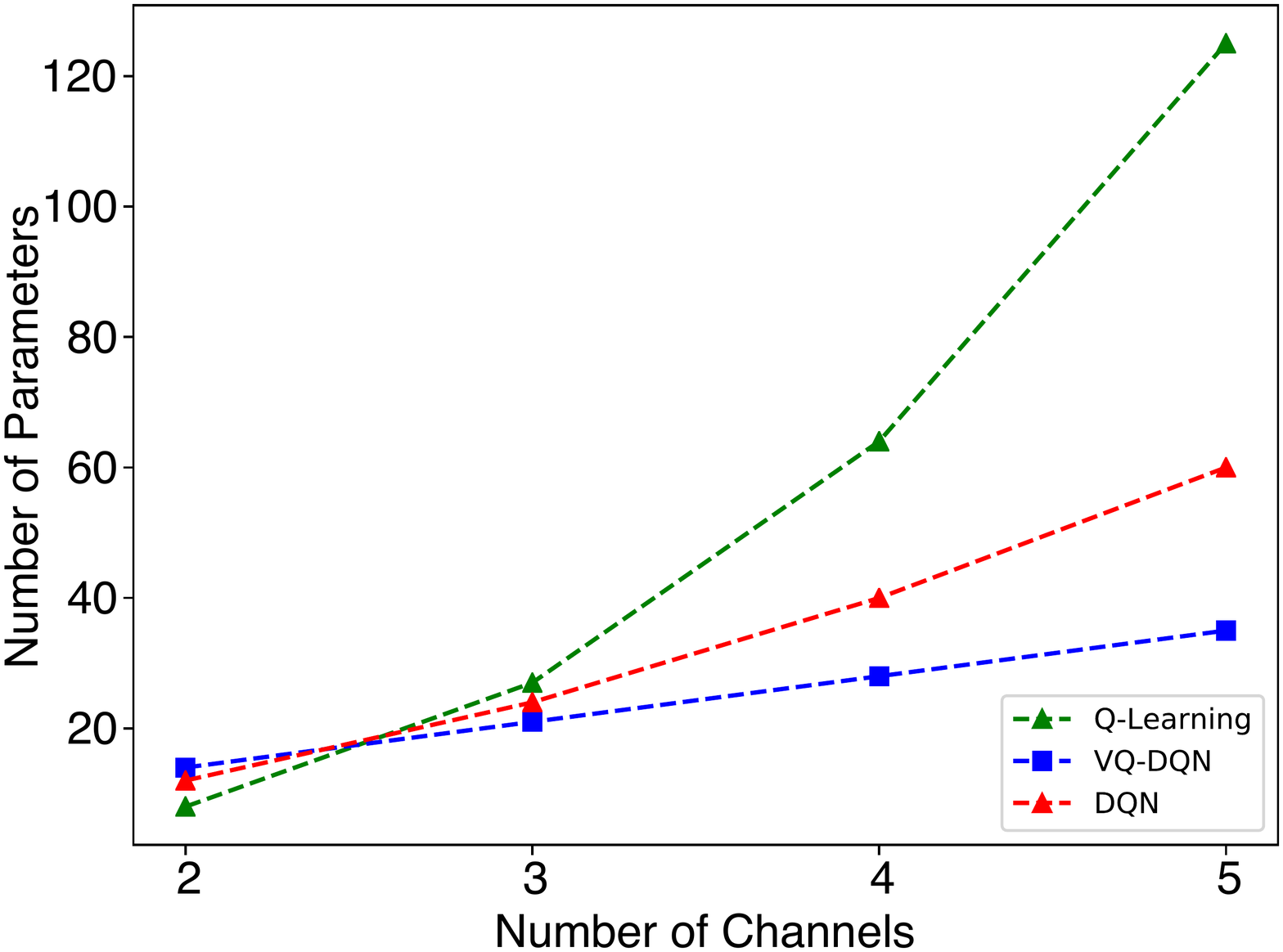}
\caption[Result:ComparisonMemoryConsumption]{{\bfseries Comparison of memory consumption in different learning schemes.} The figure shows our proposed variational quantum circuits based DQL-agent has the quantum advantage in memory consumption compared to classical $Q$-learning and DQL (DQN). Specifically, in our cognitive-radio channel selection experiment, we set up four different testing environments with 2,3,4,5 possible channels, respectively. With the classical $Q$-learning, the number of parameters grow with $n^3$, and in  DQL (DQN), the number of parameters grow with $n^2$. However, with our proposed variational quantum circuits based DQL (DQN), the number of parameters grows as $n\times(3\times2+1)$ only.}
\label{ComparisonMemoryConsumption}
\end{figure}

\textcolor{black}{Current NISQ machines are not suitable for deep quantum circuit architectures due to the lack of quantum error correction. Therefore, to utilize these near-term NISQ machines for more complex situations, it is urgent to develop quantum circuits which are not so deep in quantum gates as there will be more errors when the number of quantum gates increases. With this constraint, the proposed variational quantum circuit cannot have many parameters compared to classical neural networks. In this work, we show that for the cases we consider, variational quantum circuits with fewer iterative parameters can achieve comparable performances to classical neural networks.
}  

To compare the performance of our VQ-DQL (VQ-DQN) models with classical counterparts, we need at least one classical candidate. 
\textcolor{black}{In the frozen-lake environment, the number of parameters in our proposed VQ-DQL is $28$, while in tabular Q-Learning, the table that needs to store all the state-action information is of the size $16 \times 4 = 64$. Thus our VQ-DQL method reduces the number of parameters by $(64-28)/64 = 56.25\%$.
In the cognitive-radio environment,
the classical counterpart that we compare is from the original ns3-gym work~\cite{gawlowicz2018ns3}, and it is a fairly simple neural network consisting of a single hidden layer. From the results of Fig.~$6$ and Fig.~$8$ in that original paper, we can see that our agents converge faster than the classical ones. For example, even in the presence of noise, our agents can reach score very close to $100$ at less than $10$ iterations as shown in \figureautorefname{\ref{VQDQN_CognitiveRadio_2_3_4_5_With_IBM_NOISE}} while in the original paper~\cite{gawlowicz2018ns3}, the agents reach score $100$ in more than $20$ iterations. 
Moreover, in this example, our proposed quantum circuit uses fewer parameters than its classical counterpart.
For a given problem, there are many possible neural network architectures and it is impossible to exhaust all the possible solutions. In our case, we select the neural network architecture from the article which demonstrates the ns3 simulation framework~\cite{gawlowicz2018ns3}.
For the experiments we conduct and test in the cognitive-radio environments with $n$ possible channels, the number of parameters is $n \times (3\times2+1)$ in our variational-quantum-circuit-based DQL (DQN), while it is $2\times n^2 + 2\times n$ in the neural network based RL~\cite{gawlowicz2018ns3}.
In general, there is no guarantee that a given neural network model or a variational quantum circuit can scale as we wish when the complexity of the problem increases.
Therefore, in this work, the comparison in scaling is restricted to the cases we test. For example, in the cognitive-radio experiment, we consider only the cases with number of channels $n\leq 5$.}

In summary, the quantum advantage in our proposed method refers to the less  memory consumption, which means there are less parameters in the quantum circuits.
The quantum advantage in terms of the parameters of the quantum circuit relies on the data encoding schemes. For example, the number of parameters in an amplitude coding may be $poly(\log{} n)$ in contrast to $poly(n)$ in a standard neural network. However, it is hard to implement the amplitude encoding scheme as there is no known efficient algorithm to prepare classical vectors into quantum registers  \cite{mottonen2005transformation}.
The computational basis encoding in our proposed variational quantum  DQL (DQN) involves $n$ parameters, but there are $n^2$ parameters in the neural network based RL~\cite{gawlowicz2018ns3}, and $n^3$ in the tabular $Q$-learning, where $n$ is the dimension of input vectors.
A comparison of classical reinforcement learning algorithms with discrete action space and variational quantum deep Q networks (VQ-DQN) is given in Table~\ref{tab:qadv}.
The blue dotted lines in \figureautorefname{ \ref{ComparisonMemoryConsumption}} shows explicitly 
that the number of parameters of our proposed VQ-DQN method grows linearly with the dimension of the input vector, \textcolor{black}{at least for the cases we test, i.e.,  
 the number of channels $n\leq 5$ in the cognitive-radio experiments.}

\section{Discussions}
\label{sec:discussions}
\subsection{Overview of Quantum Machine Learning}
A general review of the field of \emph{quantum machine learning} can refer to  \cite{Dunjko2018MachineProgress,biamonte2017quantum}. As for \emph{quantum reinforcement learning}, \textcolor{black}{the early work can refer to~\cite{dong2008quantum}, and in this work, the authors use the amplitude amplification method to perform the action selection. However, such operations are hard to implement on NISQ machines. In addition, in their scheme~\cite{dong2008quantum}, there is a need to encode the environment in the quantum superposition state, which is not always possible when the agent interacts with classical environments.} 
\textcolor{black}{To study the effect of computational agents coupled to environments which are quantum-mechanical, refer to~\cite{dunjko2015framework}.}
\textcolor{black}{For a review of recent developments in quantum reinforcement learning, refer to this good review article~\cite{Dunjko2017AdvancesLearning}.}
\textcolor{black}{For example, recent work}~
\cite{Briegel2012ProjectiveIntelligence, Mautner2015ProjectiveInvestigation} have proposed a framework called \emph{projective simulation}, and the key concept in this setting is that the \emph{agent} keeps \emph{memory} of the transition history. Before executing each action, the agent will \emph{simulate} several possible outcomes according to historical data stored in the memory. It is conceptually related to the well-known Monte-Carlo Tree Search \cite{Silver2016MasteringSearch,Silver2017MasteringKnowledge} and it is interesting to investigate the quantum counterparts and possible quantum advantages. 

\vspace{-0.12cm}

\subsection{More Complex Testing Environments}

We have applied our VQ-DQN to a simple maze problem, the frozen lake environment in OpenAI Gym~\cite{Brockman2016OpenAIGym}, and to a classical spectrum control problem in cognitive radio with the ns3-gym \cite{gawlowicz2018ns3} environment.  We choose the frozen-lake environment  as an environment with low computational complexity to implement the proof-of-principle quantum DQL (DQN) experiments.
We consider the cognitive-radio problem in a wireless multi-channel environment, e.g. 802.11 networks with external interference. The objective of the agent is to select a channel free of interference in the next time slot.
We create a scalable reinforcement learning environment sim-radio-spectrum (SRS) with a customized state and an action echo in a real multi-channel spectrum scenario for the quantum DQL (DQN) demonstration of a real-world application.

Different from the benchmark environments of complex RL like Atari games,
we use simple testing environments to study the feasibility of quantum circuits-based DQL (DQN). 
Although the mainstream benchmark environments can be encoded with computational basis encoding or amplitude encoding, the number of qubits needed is intractable for numerical simulations on classical computers and also exceeds currently commercially available quantum devices. However, we could further investigate these complex RL environments with the same setting proposed in this work when large quantum machines are released.
\subsection{Scaling Up the Architecture}
\textcolor{cyan}{
  In the proposed architecture, we require the observation of the expectation value of each qubit in order to determine the next action, which can be evaluated analytically by a quantum simulation software on a classical computer. While in a real quantum computer, it is needed to perform multiple samplings of repeated measurements.
  For example, in a $n$-qubit circuit with the number of output actions to be $n$, the number of samplings in the inference stage could be a fixed number of  $1024$ as the the measurements on each qubit of the $n$ qubits can be performed simultaneously in parallel.
  Thus number of samplings needed could be pretty much fixed or may increase only slightly with the number of qubits $n$ or the number of possible output actions. Furthermore, in the proposed quantum DRL algorithm, we only care which qubit or action is the most probable with the largest expectation value and do not need the exact probability.
Thus, even though repeated experiments (measurements) are needed for our quantum DRL algorithm performed on a real quantum computer, the probabilistic decision rules of choosing the action corresponding to the highest frequency outcome with a fixed number of measurements can give better performances in terms of number of operations,
  such that the classical simulation is still not favorable when larger architectures or qubit numbers are considered.  
}%
\subsection{Simulability on a Classical Computer}
\textcolor{cyan}{
  The hardness of simulability of quantum circuits does not depend only on the number of qubits, but also on the structure of the circuit.
  In the proposed circuit architecture with fixed number of variational layers, the number of CNOT gates scales linearly with the number of qubits $n$.  Therefore, the circuit is not constant-depth. In addition, the circuit includes quantum gates beyond the Clifford group, meaning that it is not in the family of Clifford circuits. Thus, it is not obviously classically simulable when the number of qubits is large. The hardness of simulability of this model is also worth studying theoretically.
}

\subsection{Future Work: Amplitude Encoding Scheme}
Unlike the amplitude encoding, the computational basis encoding has not fully employed the quantum advantages. Although in a constraint condition of quantum simulators, we can verify the feasibility of applying quantum circuits for resolving DRL problems. The related empirical results suggest that the quantum advantages outperform both tabular $Q$-learning and neural network based RL. To obtain the ideal quantum circuits with significantly fewer parameters, one can apply amplitude encoding to reduce the complexity of parameters as small as $\mathbf{poly(\log{n})}$ in contrast to a standard neural network with $\mathbf{poly(n)}$ parameters. Future work includes an investigation of applying amplitude encoding scheme to more complex input data and variational quantum circuits to solving more sophisticated problems.

\section{Conclusions}
\label{sec:conclusions}
This is the first demonstration of variational quantum circuits to approximate the deep $Q$-value function with experience replay and target network. From the results obtained from the testing environments we consider, our proposed framework shows the quantum advantage in terms of less memory consumption and the reduction of model parameters. Specifically for the considered testing environments in this paper, the variational quantum deep $Q$-learning involves parameters as small as $\mathcal{O}(n)$, but the tabular $Q$-learning and neural network based deep $Q$-learning have $\mathcal{O}(n^3)$ and $\mathcal{O}(n^2)$ parameters, respectively.

\appendices

\section*{Acknowledgments}
\label{sec:acknowledgements}
S.Y.C. thanks Yun-Cheng Tsai for constructive discussion and great support, and thanks Cheng-Lin Hong for helping on setting computational clusters.
S.Y.C. and H.S.G. acknowledge the use of IBM Q quantum machines through the IBM Q Hub at NTU (MOST 107-2627-E-002-001-MY3). H.S.G. acknowledges and support
from the thematic group program of the National Center for Theoretical Sciences, Taiwan.



\section{Noise Information of the  IBM Q Machines Used in the Experiments}
\subsection{Device Properties for the Noisy Simulation}
\label{Noise:ibmq-poughkeepsie}
\textcolor{black}{In \sectionautorefname{\ref{sec:simulationWithNoiseFromIBM}}, we perform the numerical simulation using the IBM Qiskit simulator with the noise data downloaded from the IBM Q 20-qubit machine \texttt{ibmq-poughkeepsie} at the date 2019-07-28. The qubit relaxation time $T_1$, dephasing time $T_2$ and qubit frequency data are listed in Table~\ref{tab:ibm_noise_simulation}, the single-qubit gate error, gate length and readout error data are in Table~\ref{tab:ibm_gate_error_simulation},
  and the two-qubit coupling and corresponding CNOT gate error and gate length data are listed in Table \ref{tab:ibm_gate_coupling_error_simulation}. Note that the IBM Q system service is still under active development, meaning that the noise data will change gradually.}

 \textcolor{black}{The single-qubit gates implemented on the IBM Q systems are defined through a general single-qubit unitary gate
\begin{align}
U(\theta,\phi,\lambda) & =\left(\begin{array}{cc}
\cos(\theta/2) & -e^{i\lambda}\sin(\theta/2) \\
 -e^{i\phi}\sin(\theta/2) & e^{i\lambda+i\phi}\cos(\theta/2)
\end{array}\right),
\end{align}
with $u1(\lambda)= U(0,0,\lambda)$, $u2(\phi,\lambda)= U(\pi/2,\phi,\lambda)$, and $u3(\theta,\phi,\lambda)= U(\pi/2,\phi,\lambda)$. The single-qubit gates listed in Table~\ref{tab:ibm_gate_error_simulation} are defined as $U1=u1(\pi/2)$, $U2=u2(\pi/2,\pi/2)$,  and $U3=u3(\pi/2,\pi/2,\pi/2)$, respectively. 
In reality, the $U1$ gate is done using a frame change which means it is done in software (gate time is zero) and thus its gate error is also set to zero in Table \ref{tab:ibm_gate_error_simulation}.
}

\subsection{Device properties for the Quantum inference}
\label{Noise:ibmq-valencia}
\textcolor{black}{
The real IBM Q system we use in the experiments described in~\sectionautorefname{\ref{sec:runningOnRealQuantumComputer}} is the 5-qubit machine of ~\texttt{ibmq-valencia}.
  The qubit relaxation time $T_1$, dephasing time $T_2$ and qubit frequency of each qubit are listed in Table~\ref{tab:ibm_noise_valencia}, the single-qubit gate error, gate length and readout error data are listed in Table~\ref{tab:ibm_gate_error_valencia},
and the two-qubit coupling and corresponding CNOT gate error and gate length data are listed in
  Table~\ref{tab:ibm_coupling_gate_error_valencia}.}


\begin{table}[htbp]
\centering
\caption{\textcolor{black}{Qubit relaxation time $T_1$, dephasing time $T_2$ and qubit frequency data of the 20-qubit machine \texttt{ibmq-poughkeepsie} downloaded from the IBM Q system service at the time when the numerical simulation in~\sectionautorefname{\ref{sec:simulationWithNoiseFromIBM}} performed.}}
\label{tab:ibm_noise_simulation}
\begin{tabular}{|c|c|c|c|}
\hline
Qubit  & T1 (${\rm \mu}$s) & T2 (${\rm \mu}$s) & Frequency (GHz)\\ \hline
0  & 61.985008 & 63.458383 & 4.919976 \\ \hline
1  & 85.949810 & 117.651095 & 4.831989 \\ \hline
2  & 51.816944 & 67.602274 & 4.939780 \\ \hline
3  & 66.607063 & 67.884742 & 4.515083 \\ \hline
4  & 100.442250 & 100.323516 & 4.66276 \\ \hline
5  & 66.118221 & 65.421503 & 4.957158 \\ \hline
6  & 67.734257 & 86.704533 & 4.995404 \\ \hline
7  & 76.736639 & 17.822148 & 4.811046 \\ \hline
8  & 66.743110 & 64.9755788 & 5.013321 \\ \hline
9  & 73.998165 & 82.741817 & 5.056846 \\ \hline
10  & 68.916861 & 12.902928 & 4.719699 \\ \hline
11  & 68.614596 & 86.433312 & 4.900151 \\ \hline
12  & 45.072575 & 9.751853 & 4.772410 \\ \hline
13  & 57.462386 & 23.218794 & 5.110509 \\ \hline
14  & 72.111045 & 83.851059 & 4.990636 \\ \hline
15  & 92.118025 & 41.604544 & 4.806460 \\ \hline
16  & 89.207166 & 61.459798 & 4.955924 \\ \hline
17  & 106.508278 & 18.273605 & 4.599708 \\ \hline
18  & 93.291790 & 94.709020 & 4.828349 \\ \hline
19  & 77.475090 & 99.860137 & 4.938456 \\ \hline
\end{tabular}
\end{table}

\begin{table}[htbp]
\centering
\caption{\textcolor{black}{
Two-qubit coupling and CNOT gate error and gate length data of the 5-qubit machine \texttt{ibmq-valencia}  downloaded from the IBM Q system service
 at the time when the experiment in~\sectionautorefname{\ref{sec:runningOnRealQuantumComputer}} performed.}}
\label{tab:ibm_coupling_gate_error_valencia}
\begin{tabular}{|c|c|c|c|}
\hline
Index  & Coupling pair & CNOT gate error & CNOT gate length (ns) \\ \hline
0 &  [0, 1] &      0.008928 & 320.0000 \\ \hline
1 &  [1, 0] &      0.008928 & 284.4444 \\ \hline
2 &  [1, 2] &      0.007558 & 341.3333 \\ \hline
3 &  [1, 3] &      0.011670 & 661.3333 \\ \hline
4 &  [2, 1] &      0.007558 & 376.8889 \\ \hline
5 &  [3, 1] &      0.011670 & 696.8889 \\ \hline
6 &  [3, 4] &      0.009659 & 305.7778 \\ \hline
7 &  [4, 3] &      0.009659 & 341.3333 \\ \hline
\end{tabular}
\end{table}
\begin{table}[htbp]
\centering
\caption{\textcolor{black}{Two-qubit coupling and CNOT gate error and gate length data of the 20-qubit machine \texttt{ibmq-poughkeepsie}  downloaded from the IBM Q system service at the time when the numerical simulation in~\sectionautorefname{\ref{sec:simulationWithNoiseFromIBM}} performed. The two-qubit CNOT gate error and lengths are usually different for different two-qubit coupling pairs due to different coupling strength and different qubit frequencies in the qubit pairs.}}
\label{tab:ibm_gate_coupling_error_simulation}
\begin{tabular}{|c|c|c|c|}
\hline
Index  &   Coupling pair & CNOT gate error & CNOT gate length (ns)\\ \hline
0 &     [0, 1] &      0.019174 & 455.1111 \\ \hline
1 &    [0, 5] &       0.022823 & 433.7778 \\ \hline
2 &    [1, 0] &       0.019174 & 455.1111 \\ \hline
3 &    [1, 2] &       0.018289 & 455.1111 \\ \hline
4 &    [2, 1] &       0.018289 & 455.1111 \\ \hline
5 &    [2, 3] &       0.027451 & 839.1111 \\ \hline
6 &    [3, 2] &      0.027451 & 839.1111 \\ \hline
7 &    [3, 4] &      0.023150 & 832.0000 \\ \hline
8 &    [4, 3] &      0.023150 & 832.0000 \\ \hline
9 &    [4, 9] &      0.123852 & 888.8889 \\ \hline
10 &   [5, 0] &      0.022823 & 433.7778 \\ \hline
11 &   [5, 6] &      0.017515 & 455.1111 \\ \hline
12 &  [5, 10] &      0.022059 & 476.4444 \\ \hline
13 &   [6, 5] &      0.017515 & 455.1111 \\ \hline
14 &   [6, 7] &      0.029664 & 618.6667 \\ \hline
15 &   [7, 6] &      0.029664 & 618.6667 \\ \hline
16 &   [7, 8] &      0.024846 & 398.2222 \\ \hline
17 &  [7, 12] &      0.024576 & 554.6667 \\ \hline
18 &   [8, 7] &      0.024846 & 398.2222 \\ \hline
19 &   [8, 9] &      0.021136 & 426.6667 \\ \hline
20 &   [9, 4] &      0.123852 & 888.8889 \\ \hline
21 &   [9, 8] &      0.021136 & 426.6667 \\ \hline
22 &  [9, 14] &      0.032180 & 568.8889 \\ \hline
23 &  [10, 5] &      0.022059 & 476.4444 \\ \hline
24 & [10, 11] &      0.020104 & 661.3333 \\ \hline
25 & [10, 15] &      0.014944 & 476.4444 \\ \hline
26 & [11, 10] &      0.020104 & 661.3333 \\ \hline
27 & [11, 12] &      0.016619 & 448.0000 \\ \hline
28 &  [12, 7] &      0.024576 & 554.6667 \\ \hline
29 & [12, 11] &      0.016619 & 448.0000 \\ \hline
30 & [12, 13] &      0.132743 & 618.6667 \\ \hline
31 & [13, 12] &      0.132743 & 618.6667 \\ \hline
32 & [13, 14] &      0.021946 & 398.2222 \\ \hline
33 &  [14, 9] &      0.032180 & 568.8889 \\ \hline
34 & [14, 13] &      0.021946 & 398.2222 \\ \hline
35 & [14, 19] &      0.017132 & 469.3333 \\ \hline
36 & [15, 10] &      0.014944 & 476.4444 \\ \hline
37 & [15, 16] &      0.021673 & 789.3333 \\ \hline
38 & [16, 15] &      0.021673 & 789.3333 \\ \hline
39 & [16, 17] &      0.019825 & 704.0000 \\ \hline
40 & [17, 16] &      0.019825 & 704.0000 \\ \hline
41 & [17, 18] &      0.020812 & 704.0000 \\ \hline
42 & [18, 17] &      0.020812 & 704.0000 \\ \hline
43 & [18, 19] &      0.015828 & 419.5556 \\ \hline
44 & [19, 14] &      0.017132 & 469.3333 \\ \hline
45 & [19, 18] &      0.015828 & 419.5556 \\ \hline
\end{tabular}
\end{table}
%


%

\begin{table*}[htbp]
\centering
\caption{\textcolor{black}{
    Single-qubit U1, U2 and U3 gate errors, and readout error data
of the 20-qubit  machine \texttt{ibmq-poughkeepsie} downloaded from the IBM Q system service at the time when the numerical simulation in~\sectionautorefname{\ref{sec:simulationWithNoiseFromIBM}} performed.
Different single-qubit gates have different gate lengths, but the gate length of the same single-qubit gate is the same for every qubit. The gate lengths of the Identity (Id) gate, U1 gate, U2 gate, and U3 gate, are
    113.7778 ns,  0.0 ns,    103.1111 ns, and  206.2222 ns,  respectively.}}
\label{tab:ibm_gate_error_simulation}
\begin{tabular}{|c|c|c|c|c|c|}
\hline
Qubit & Id gate error & U1 gate error & U2 gate error & U3 gate error & Readout error \\ \hline
0  &      0.001207 &           0.0 &      0.001207 &      0.002413 &         0.032 \\ \hline
1  &      0.000927 &           0.0 &      0.000927 &      0.001854 &         0.031 \\ \hline
2  &      0.001200 &           0.0 &      0.001200 &      0.002401 &         0.030 \\ \hline
3  &      0.001287 &           0.0 &      0.001287 &      0.002574 &         0.043 \\ \hline
4  &      0.001383 &           0.0 &      0.001383 &      0.002767 &         0.067 \\ \hline
5  &      0.001075 &           0.0 &      0.001075 &      0.002150 &         0.031 \\ \hline
6  &      0.001545 &           0.0 &      0.001545 &      0.003091 &         0.040 \\ \hline
7  &      0.003174 &           0.0 &      0.003174 &      0.006349 &         0.070 \\ \hline
8  &      0.001416 &           0.0 &      0.001416 &      0.002832 &         0.031 \\ \hline
9  &      0.001986 &           0.0 &      0.001986 &      0.003972 &         0.028 \\ \hline
10 &      0.002024 &           0.0 &      0.002024 &      0.004047 &         0.061 \\ \hline
11 &      0.000570 &           0.0 &      0.000570 &      0.001140 &         0.025 \\ \hline
12 &      0.002100 &           0.0 &      0.002100 &      0.004200 &         0.059 \\ \hline
13 &      0.003205 &           0.0 &      0.003205 &      0.006410 &         0.025 \\ \hline
14 &      0.001408 &           0.0 &      0.001408 &      0.002816 &         0.029 \\ \hline
15 &      0.000900 &           0.0 &      0.000900 &      0.001800 &         0.023 \\ \hline
16 &      0.000901 &           0.0 &      0.000901 &      0.001801 &         0.018 \\ \hline
17 &      0.001404 &           0.0 &      0.001404 &      0.002807 &         0.080 \\ \hline
18 &      0.000769 &           0.0 &      0.000769 &      0.001538 &         0.023 \\ \hline
19 &      0.000943 &           0.0 &      0.000943 &      0.001886 &         0.023 \\ \hline
\end{tabular}
\end{table*}

\begin{table*}[htbp]
\centering
\caption{\textcolor{black}{
Single-qubit U1, U2 and U3 gate errors, and readout error data
of the 20-qubit machine \texttt{ibmq-valencia}  downloaded from the IBM Q system service  at the time when the experiment in~\sectionautorefname{\ref{sec:runningOnRealQuantumComputer}} performed. The gate lengths of the Identity (Id) gate, U1 gate, U2 gate, and U3 gate, are
    35.55556 ns,  0.0 ns,    35.55556 ns, and  71.11111 ns,  respectively.}}
\label{tab:ibm_gate_error_valencia}
\begin{tabular}{|c|c|c|c|c|c|}
\hline
Qubit   & Id gate error & U1 gate error & U2 gate error & U3 gate error & Readout error \\ \hline
0 &      0.000361 &           0.0 &      0.000361 &      0.000721 &       0.01750 \\ \hline
1 &      0.000327 &           0.0 &      0.000327 &      0.000654 &       0.01500 \\ \hline
2 &      0.001065 &           0.0 &      0.001065 &      0.002129 &       0.12250 \\ \hline
3 &      0.000295 &           0.0 &      0.000295 &      0.000590 &       0.01500 \\ \hline
4 &      0.000294 &           0.0 &      0.000294 &      0.000588 &       0.01375 \\ \hline
\end{tabular}
\end{table*}

\begin{table}[htbp]
\centering
\caption{\textcolor{black}{
Qubit relaxation time $T_1$, dephasing time $T_2$ and qubit frequency data of the 5-qubit machine \texttt{ibmq-valencia} downloaded from the IBM Q system service
   at the time when the experiment in~\sectionautorefname{\ref{sec:runningOnRealQuantumComputer}} performed.}}
\label{tab:ibm_noise_valencia}
\begin{tabular}{|c|c|c|c|}
\hline
 Qubit &    T1 (${\rm \mu}$s)     &  T2 (${\rm \mu}$s)       &   Frequency (GHz)    \\ \hline
0 & 119.527115 &  86.931245 &   4.744506 \\ \hline
1 & 112.397617 &  83.546189 &   4.650691 \\ \hline
2 & 123.174284 &  16.005230 &   4.792272 \\ \hline
3 & 135.673765 &  66.236028 &   4.834118 \\ \hline
4 &  97.472783 & 103.480782 &   4.959371 \\ \hline
\end{tabular}
\end{table}
%






\clearpage
\bibliographystyle{ieeetr}
\bibliography{FINALVERSION.bib}

\begin{thebibliography}{10}

\bibitem{lecun2015deep}
Y.~LeCun, Y.~Bengio, and G.~Hinton, ``Deep learning,'' {\em Nature}, vol.~521,
  pp.~436--444, May 2015.

\bibitem{Simonyan2014VeryRecognition}
K.~Simonyan and A.~Zisserman, ``Very deep convolutional networks for
  large-scale image recognition,'' in {\em ICLR}, 2015.

\bibitem{Szegedy2014GoingConvolutions}
C.~Szegedy, W.~Liu, Y.~Jia, P.~Sermanet, S.~Reed, D.~Anguelov, D.~Erhan,
  V.~Vanhoucke, and A.~Rabinovich, ``Going deeper with convolutions,'' in {\em
  2015 IEEE Conference on Computer Vision and Pattern Recognition (CVPR)},
  pp.~1--9, IEEE, June 2015.

\bibitem{Voulodimos2018DeepReview}
A.~Voulodimos, N.~Doulamis, A.~Doulamis, and E.~Protopapadakis, ``Deep learning
  for computer vision: {A} brief review,'' {\em Comput. Intell. Neurosci.},
  vol.~2018, pp.~1--13, 2018.

\bibitem{Sutskever2014SequenceNetworks}
I.~Sutskever, O.~Vinyals, and Q.~V. Le, ``Sequence to sequence learning with
  neural networks,'' in {\em Adv. Neural Inf. Process. Syst.}, pp.~3104--3112,
  2014.

\bibitem{kao2019reinforcement}
S.-C. Kao, C.-H.~H. Yang, P.-Y. Chen, X.~Ma, and T.~Krishna, ``Reinforcement
  learning based interconnection routing for adaptive traffic optimization,''
  in {\em Proceedings of the 13th IEEE/ACM International Symposium on
  Networks-on-Chip}, pp.~1--2, 2019.

\bibitem{Silver2016MasteringSearch}
D.~Silver, A.~Huang, C.~J. Maddison, A.~Guez, L.~Sifre, G.~van~den Driessche,
  J.~Schrittwieser, I.~Antonoglou, V.~Panneershelvam, M.~Lanctot, S.~Dieleman,
  D.~Grewe, J.~Nham, N.~Kalchbrenner, I.~Sutskever, T.~Lillicrap, M.~Leach,
  K.~Kavukcuoglu, T.~Graepel, and D.~Hassabis, ``Mastering the game of go with
  deep neural networks and tree search,'' {\em Nature}, vol.~529, pp.~484--489,
  Jan. 2016.

\bibitem{Borin2019ApproximatingStates}
A.~Borin and D.~A. Abanin, ``Approximating power of machine-learning ansatz for
  quantum many-body states,'' {\em arXiv preprint arXiv:1901.08615}, 2019.

\bibitem{Carleo2019NetKet:Systems}
G.~Carleo, K.~Choo, D.~Hofmann, J.~E. Smith, T.~Westerhout, F.~Alet, E.~J.
  Davis, S.~Efthymiou, I.~Glasser, S.-H. Lin, M.~Mauri, G.~Mazzola, C.~B.
  Mendl, E.~van Nieuwenburg, O.~O’Reilly, H.~Théveniaut, G.~Torlai,
  F.~Vicentini, and A.~Wietek, ``{NetKet:} {A} machine learning toolkit for
  many-body quantum systems,'' {\em SoftwareX}, vol.~10, p.~100311, July 2019.

\bibitem{Carleo2019MachineSciences}
G.~Carleo, I.~Cirac, K.~Cranmer, L.~Daudet, M.~Schuld, N.~Tishby,
  L.~Vogt-Maranto, and L.~Zdeborová, ``Machine learning and the physical
  sciences,'' {\em Rev. Mod. Phys.}, vol.~91, p.~045002, Dec 2019.

\bibitem{Canabarro2019UnveilingLearning}
A.~Canabarro, F.~F. Fanchini, A.~L. Malvezzi, R.~Pereira, and R.~Chaves,
  ``Unveiling phase transitions with machine learning,'' {\em Phys. Rev. B},
  vol.~100, p.~045129, July 2019.

\bibitem{An2019DeepControl}
Z.~An and D.~Zhou, ``Deep reinforcement learning for quantum gate control,''
  {\em Europhys. Lett.}, vol.~126, p.~60002, July 2019.

\bibitem{Flurin2018UsingObservations}
E.~Flurin, L.~Martin, S.~Hacohen-Gourgy, and I.~Siddiqi, ``Using a recurrent
  neural network to reconstruct quantum dynamics of a superconducting qubit
  from physical observations,'' {\em Phys. Rev. X}, vol.~10, p.~011006, Jan
  2020.

\bibitem{Andreasson2018QuantumLearning}
P.~Andreasson, J.~Johansson, S.~Liljestrand, and M.~Granath, ``Quantum error
  correction for the toric code using deep reinforcement learning,'' {\em
  Quantum}, vol.~3, p.~183, Sept. 2019.

\bibitem{Nautrup2018OptimizingLearning}
H.~Poulsen~Nautrup, N.~Delfosse, V.~Dunjko, H.~J. Briegel, and N.~Friis,
  ``Optimizing quantum error correction codes with reinforcement learning,''
  {\em Quantum}, vol.~3, p.~215, Dec. 2019.

\bibitem{cross2018ibm}
A.~Cross, ``The {IBM} q experience and {QISKit} open-source quantum computing
  software,'' in {\em APS Meeting Abstracts}, 2018.

\bibitem{lanting2014entanglement}
T.~Lanting, A.~Przybysz, A.~Smirnov, F.~Spedalieri, M.~Amin, A.~Berkley,
  R.~Harris, F.~Altomare, S.~Boixo, P.~Bunyk, N.~Dickson, C.~Enderud,
  J.~Hilton, E.~Hoskinson, M.~Johnson, E.~Ladizinsky, N.~Ladizinsky,
  R.~Neufeld, T.~Oh, I.~Perminov, C.~Rich, M.~Thom, E.~Tolkacheva, S.~Uchaikin,
  A.~Wilson, and G.~Rose, ``Entanglement in a quantum annealing processor,''
  {\em Phys. Rev. X}, vol.~4, p.~021041, May 2014.

\bibitem{gottesman1997stabilizer}
D.~Gottesman, ``Stabilizer codes and quantum error correction,'' {\em arXiv
  preprint quant-ph/9705052}, 1997.

\bibitem{gottesman1998theory}
D.~Gottesman, ``Theory of fault-tolerant quantum computation,'' {\em Phys. Rev.
  A}, vol.~57, pp.~127--137, Jan. 1998.

\bibitem{Mitarai2018QuantumLearning}
K.~Mitarai, M.~Negoro, M.~Kitagawa, and K.~Fujii, ``Quantum circuit learning,''
  {\em Phys. Rev. A}, vol.~98, p.~032309, Sept. 2018.

\bibitem{du2018expressive}
Y.~Du, M.-H. Hsieh, T.~Liu, and D.~Tao, ``The expressive power of parameterized
  quantum circuits,'' {\em arXiv preprint arXiv:1810.11922}, 2018.

\bibitem{Preskill2018quantumcomputingin}
J.~Preskill, ``Quantum computing in the {NISQ} era and beyond,'' {\em Quantum},
  vol.~2, p.~79, Aug. 2018.

\bibitem{Schuld2018Circuit-centricClassifiers}
M.~Schuld, A.~Bocharov, K.~Svore, and N.~Wiebe, ``Circuit-centric quantum
  classifiers,'' {\em arXiv preprint arXiv:1804.00633}, 2018.

\bibitem{havlivcek2019supervised}
V.~Havlíček, A.~D. Córcoles, K.~Temme, A.~W. Harrow, A.~Kandala, J.~M. Chow,
  and J.~M. Gambetta, ``Supervised learning with quantum-enhanced feature
  spaces,'' {\em Nature}, vol.~567, pp.~209--212, Mar. 2019.

\bibitem{goodfellow2014generative}
I.~Goodfellow, J.~Pouget-Abadie, M.~Mirza, B.~Xu, D.~Warde-Farley, S.~Ozair,
  A.~Courville, and Y.~Bengio, ``Generative adversarial nets,'' in {\em Adv.
  Neural Inf. Process. Syst.}, pp.~2672--2680, 2014.

\bibitem{SuttonReinforcementIntroduction}
R.~S. Sutton and A.~G. Barto, {\em Reinforcement Learning}.
\newblock Springer US, 1992.

\bibitem{Mnih2016AsynchronousLearning}
V.~Mnih, A.~P. Badia, M.~Mirza, A.~Graves, T.~Lillicrap, T.~Harley, D.~Silver,
  and K.~Kavukcuoglu, ``Asynchronous methods for deep reinforcement learning,''
  in {\em ICML}, pp.~1928--1937, 2016.

\bibitem{Mnih2015Human-levelLearning}
V.~Mnih, K.~Kavukcuoglu, D.~Silver, A.~A. Rusu, J.~Veness, M.~G. Bellemare,
  A.~Graves, M.~Riedmiller, A.~K. Fidjeland, G.~Ostrovski, S.~Petersen,
  C.~Beattie, A.~Sadik, I.~Antonoglou, H.~King, D.~Kumaran, D.~Wierstra,
  S.~Legg, and D.~Hassabis, ``Human-level control through deep reinforcement
  learning,'' {\em Nature}, vol.~518, pp.~529--533, Feb. 2015.

\bibitem{Brockman2016OpenAIGym}
G.~Brockman, V.~Cheung, L.~Pettersson, J.~Schneider, J.~Schulman, J.~Tang, and
  W.~Zaremba, ``Openai gym,'' {\em arXiv preprint arXiv:1606.01540}, 2016.

\bibitem{gawlowicz2018ns3}
P.~Gawłowicz and A.~Zubow, ``ns-3 meets {OpenAI} gym,'' in {\em Proceedings of
  the 22nd International ACM Conference on Modeling, Analysis and Simulation of
  Wireless and Mobile Systems - MSWIM '19}, ACM Press, Nov. 2019.

\bibitem{kandala2017hardware}
A.~Kandala, A.~Mezzacapo, K.~Temme, M.~Takita, M.~Brink, J.~M. Chow, and J.~M.
  Gambetta, ``Hardware-efficient variational quantum eigensolver for small
  molecules and quantum magnets,'' {\em Nature}, vol.~549, pp.~242--246, Sept.
  2017.

\bibitem{farhi2014quantum}
E.~Farhi, J.~Goldstone, and S.~Gutmann, ``A quantum approximate optimization
  algorithm,'' {\em arXiv preprint arXiv:1411.4028}, 2014.

\bibitem{mcclean2016theory}
J.~R. McClean, J.~Romero, R.~Babbush, and A.~Aspuru-Guzik, ``The theory of
  variational hybrid quantum-classical algorithms,'' {\em New J. Phys.},
  vol.~18, p.~023023, Feb. 2016.

\bibitem{Farhi2018ClassificationProcessors}
E.~Farhi and H.~Neven, ``Classification with quantum neural networks on near
  term processors,'' {\em arXiv preprint arXiv:1802.06002}, 2018.

\bibitem{arute2019quantum}
F.~Arute, K.~Arya, R.~Babbush, D.~Bacon, J.~C. Bardin, R.~Barends, R.~Biswas,
  S.~Boixo, F.~G. Brandao, D.~A. Buell, {\em et~al.}, ``Quantum supremacy using
  a programmable superconducting processor,'' {\em Nature}, vol.~574, no.~7779,
  pp.~505--510, 2019.

\bibitem{harrow2017quantum}
A.~W. Harrow and A.~Montanaro, ``Quantum computational supremacy,'' {\em
  Nature}, vol.~549, no.~7671, pp.~203--209, 2017.

\bibitem{Schuld2019EvaluatingHardware}
M.~Schuld, V.~Bergholm, C.~Gogolin, J.~Izaac, and N.~Killoran, ``Evaluating
  analytic gradients on quantum hardware,'' {\em Phys. Rev. A}, vol.~99,
  p.~032331, Mar. 2019.

\bibitem{Schuld2018InformationEncoding}
M.~Schuld and F.~Petruccione, {\em Information Encoding}, ch.~Information
  Encoding, pp.~139--171.
\newblock Cham: Springer International Publishing, 2018.

\bibitem{hornik1989multilayer}
K.~Hornik, M.~Stinchcombe, H.~White, {\em et~al.}, ``Multilayer feedforward
  networks are universal approximators.,'' {\em Neural networks}, vol.~2,
  no.~5, pp.~359--366, 1989.

\bibitem{sim2019expressibility}
S.~Sim, P.~D. Johnson, and A.~Aspuru‐Guzik, ``Expressibility and entangling
  capability of parameterized quantum circuits for hybrid quantum-classical
  algorithms,'' {\em Adv Quantum Tech}, vol.~2, p.~1900070, Oct. 2019.

\bibitem{9053342}
C.~H. {Yang}, J.~{Qi}, P.~{Chen}, Y.~{Ouyang}, I.~D. {Hung}, C.~{Lee}, and
  X.~{Ma}, ``Enhanced adversarial strategically-timed attacks against deep
  reinforcement learning,'' in {\em ICASSP 2020 - 2020 IEEE International
  Conference on Acoustics, Speech and Signal Processing (ICASSP)},
  pp.~3407--3411, 2020.

\bibitem{Bergholm2018PennyLane:Computations}
V.~Bergholm, J.~Izaac, M.~Schuld, C.~Gogolin, and N.~Killoran, ``Pennylane:
  {Automatic} differentiation of hybrid quantum-classical computations,'' {\em
  arXiv preprint arXiv:1811.04968}, 2018.

\bibitem{Paszke2017AutomaticPyTorch}
S.~Zagoruyko, A.~Lerer, T.-Y. Lin, P.~Pinheiro, S.~Gross, S.~Chintala, and
  P.~Dollar, ``A {MultiPath} network for object detection,'' in {\em Procedings
  of the British Machine Vision Conference 2016}, British Machine Vision
  Association, 2016.

\bibitem{Tieleman2012}
T.~Tieleman and G.~Hinton, ``{Lecture 6.5---{RmsProp:} {Divide} the gradient by
  a running average of its recent magnitude}.'' COURSERA: Neural Networks for
  Machine Learning, 2012.

\bibitem{mottonen2005transformation}
M.~M{\"o}tt{\"o}nen, J.~J. Vartiainen, V.~Bergholm, and M.~M. Salomaa,
  ``Transformation of quantum states using uniformly controlled rotations,''
  {\em Quant. Inf. Comp.}, vol.~5, no.~6, pp.~467--473, 2005.

\bibitem{Dunjko2018MachineProgress}
V.~Dunjko and H.~J. Briegel, ``Machine learning \& artificial intelligence in
  the quantum domain: {A} review of recent progress,'' {\em Rep. Prog. Phys.},
  vol.~81, p.~074001, June 2018.

\bibitem{biamonte2017quantum}
J.~Biamonte, P.~Wittek, N.~Pancotti, P.~Rebentrost, N.~Wiebe, and S.~Lloyd,
  ``Quantum machine learning,'' {\em Nature}, vol.~549, pp.~195--202, Sept.
  2017.

\bibitem{dong2008quantum}
D.~Dong, C.~Chen, H.~Li, and T.-J. Tarn, ``Quantum reinforcement learning,''
  {\em IEEE Transactions on Systems, Man, and Cybernetics, Part B
  (Cybernetics)}, vol.~38, no.~5, pp.~1207--1220, 2008.

\bibitem{dunjko2015framework}
V.~Dunjko, J.~M. Taylor, and H.~J. Briegel, ``Framework for learning agents in
  quantum environments,'' {\em arXiv preprint arXiv:1507.08482}, 2015.

\bibitem{Dunjko2017AdvancesLearning}
V.~Dunjko, J.~M. Taylor, and H.~J. Briegel, ``Advances in quantum reinforcement
  learning,'' in {\em 2017 IEEE International Conference on Systems, Man, and
  Cybernetics (SMC)}, pp.~282--287, IEEE, Oct. 2017.

\bibitem{Briegel2012ProjectiveIntelligence}
H.~J. Briegel and G.~De~las Cuevas, ``Projective simulation for artificial
  intelligence,'' {\em Sci Rep}, vol.~2, p.~400, May 2012.

\bibitem{Mautner2015ProjectiveInvestigation}
J.~Mautner, A.~Makmal, D.~Manzano, M.~Tiersch, and H.~J. Briegel, ``Projective
  simulation for classical learning agents: {A} comprehensive investigation,''
  {\em New Gener. Comput.}, vol.~33, pp.~69--114, Jan. 2015.

\bibitem{Silver2017MasteringKnowledge}
D.~Silver, J.~Schrittwieser, K.~Simonyan, I.~Antonoglou, A.~Huang, A.~Guez,
  T.~Hubert, L.~Baker, M.~Lai, A.~Bolton, Y.~Chen, T.~Lillicrap, F.~Hui,
  L.~Sifre, G.~van~den Driessche, T.~Graepel, and D.~Hassabis, ``Mastering the
  game of go without human knowledge,'' {\em Nature}, vol.~550, pp.~354--359,
  Oct. 2017.

\end{thebibliography}


\clearpage
\begin{IEEEbiography}[{\includegraphics[width=1in,height=1.25in,clip,keepaspectratio]{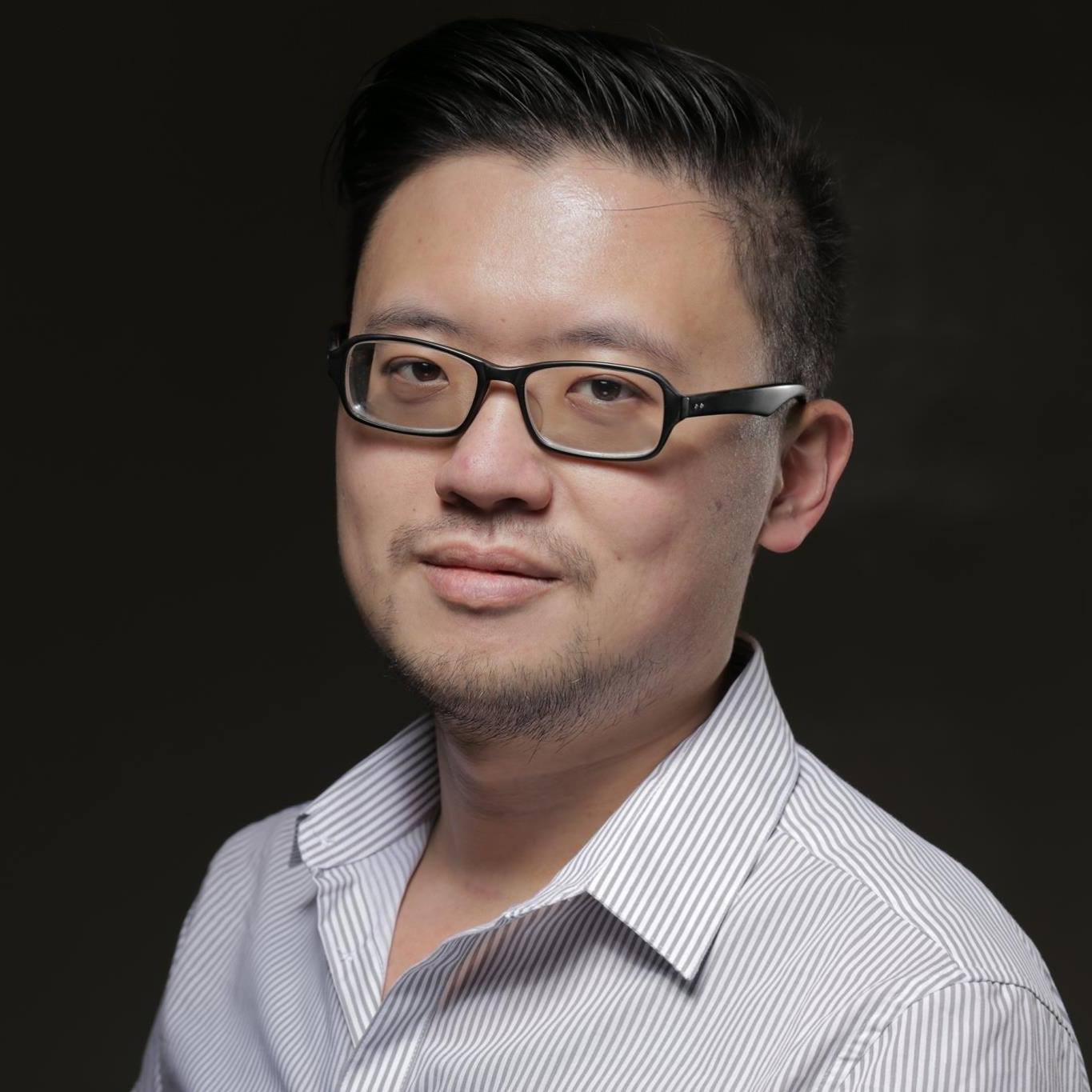}}]{Samuel Yen-Chi Chen} received the B.S. degree in physics and M.D. degree in medicine from National Taiwan University, Taipei, Taiwan, in 2016.
He is currently pursuing the Ph.D. degree in physics from National Taiwan 
University, Taipei, Taiwan. His research focus on combining quantum computing and machine learning. He was the recipient of Theoretical High-Energy Physics Fellowship from Chen Cheng Foundation in 2014, Theoretical Physics Fellowship from National Taiwan University Center for Theoretical Physics in 2015 and First Prize in the Software Competition (Research Category) from Xanadu Quantum Technologies in 2019.
\end{IEEEbiography}

\begin{IEEEbiography}[{\includegraphics[width=1in,height=1.25in,clip,keepaspectratio]{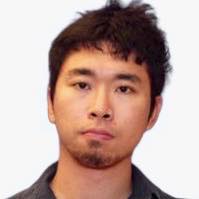}}]{Chao-Han Huck Yang} is currently working towards the Ph.D. degree in the school of electrical and computer engineering at the Georgia Institute of Technology, Atlanta, GA, USA. He received a B.S. degree from National Taiwan University, Taipei, Taiwan, in 2016. His recent research interests focus on adversarial robustness of deep neural networks and reinforcement learning with real-world applications on speech processing, communication networks, and audio-visual processing. He is a student member of IEEE society and the recipient of the \emph{Wallace H. Coulter Fellowship} from Georgia Institute of Technology in \emph{2017-18}. He received IEEE SPS travel grant for \emph{ICIP 2019}, 1st Prize on the Research Track of Xanadu Quantum Software Global Competition, and DeepMind travel award for \emph{NeurIPS} 2019. He took research interns in the Image and Visual Representation Lab (IVRL), Ecole Polytechnique Federale de Lausanne (EPFL), Switzerland, in 2018, KAUST, and Amazon Alexa Research.
\end{IEEEbiography}

\begin{IEEEbiography}[{\includegraphics[width=1in,height=1.25in,clip,keepaspectratio]{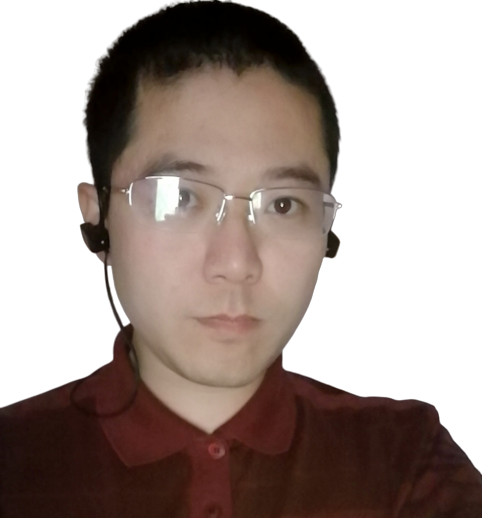}}]{Jun Qi} (S'10--M'16) is currently a Ph.D. candidate in the School of Electrical and Computer Engineering at Georgia Institute of Technology. Previously, he was a researcher at Microsoft Research during the year 2017 after he obtained two Masters of Electrical Engineering from the University of Washington, Seattle, WA, USA in 2016 and Tsinghua University in 2013, respectively. His research focuses on non-convex optimization and statistical learning for understanding deep learning systems, tensor decomposition and applications in machine learning, deep learning, submodular optimization, and speech processing.
\end{IEEEbiography}

\begin{IEEEbiography}[{\includegraphics[width=1in,height=1.25in,clip,keepaspectratio]{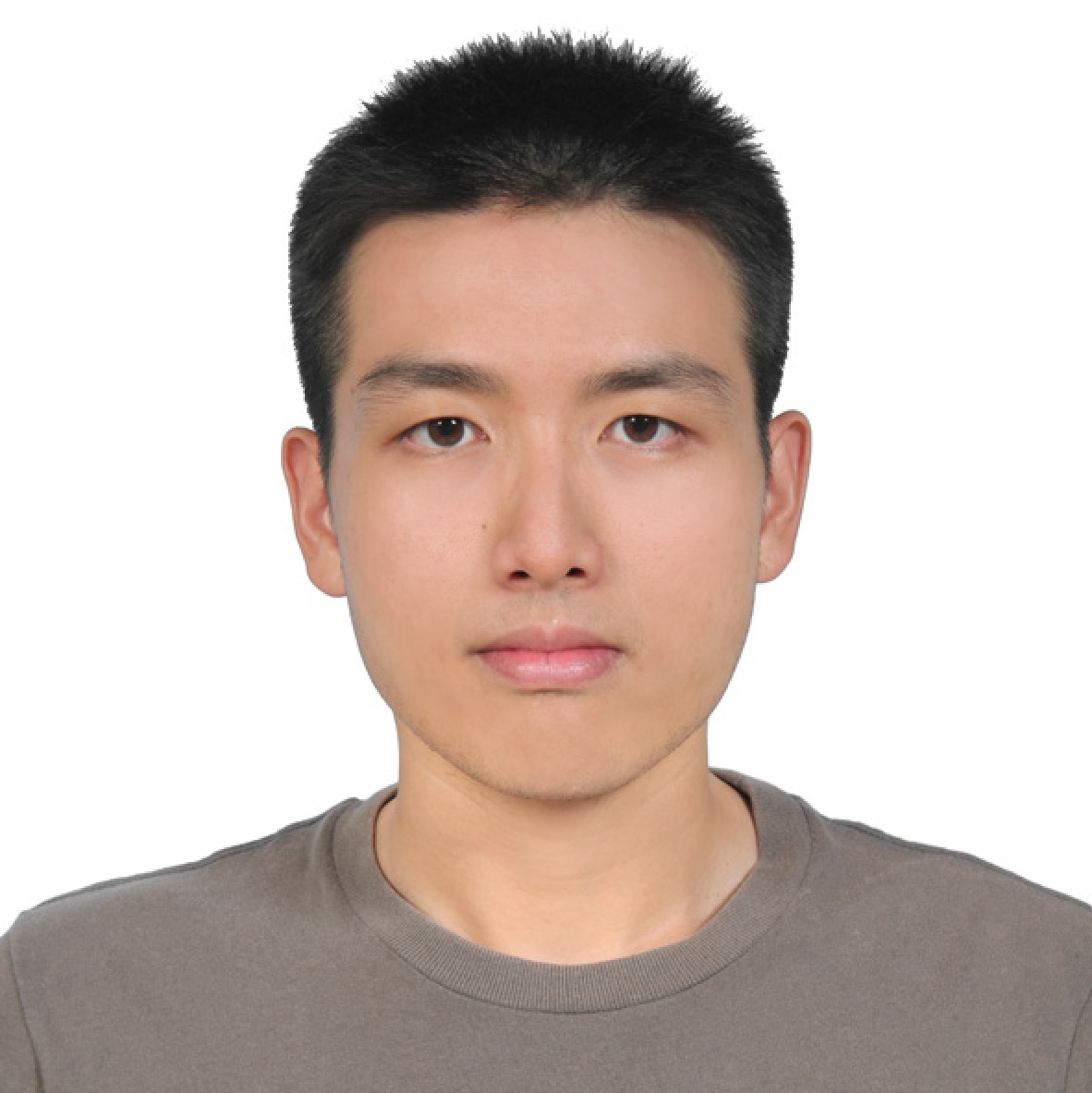}}]{Pin-Yu Chen} (S'10--M'16) is currently a research staff member at IBM Thomas J. Watson Research Center, Yorktown Heights, NY, USA. He is also the chief scientist of RPI-IBM AI Research Collaboration and PI of ongoing MIT-IBM Watson AI Lab projects. Dr. Chen received his Ph.D. degree in electrical engineering and computer science and M.A. degree in Statistics from the University of Michigan, Ann Arbor, USA, in 2016. He received his M.S. degree in communication engineering from National Taiwan University, Taiwan, in 2011 and B.S. degree in electrical engineering and computer science (undergraduate honors program) from National Chiao Tung University, Taiwan, in 2009. 
Dr. Chen’s recent research is on adversarial machine learning and robustness of neural networks. His long-term research vision is building trustworthy machine learning systems. He has published more than 20 papers on trustworthy machine learning at major AI and machine learning conferences, given tutorials at CVPR’20, ECCV’20, ICASSP’20, KDD’19 and Big Data’18, and co-organized several workshops on adversarial learning for machine learning and data mining. His research interest also includes graph and network data analytics and their applications to data mining, machine learning, signal processing, and cyber security. He was the recipient of the Chia-Lun Lo Fellowship from the University of Michigan Ann Arbor. He received the NIPS 2017 Best Reviewer Award, and was also the recipient of the IEEE GLOBECOM 2010 GOLD Best Paper Award. Dr. Chen is currently on the editorial board of PLOS ONE. 
At IBM Research, Dr. Chen has co-invented more than 20 U.S. patents. In 2019, he received two Outstanding Research Accomplishments on research in adversarial robustness and trusted AI, and one Research Accomplishment on research in graph learning and analysis.
\end{IEEEbiography}

\begin{IEEEbiography}[{\includegraphics[width=1in,height=1.25in,clip,keepaspectratio]{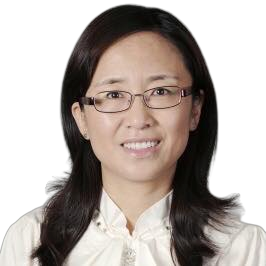}}]{Xiaoli Ma} (S'10--M'16) is Professor in School of Electrical and Computer Engineering at Georgia Institute of Technology, Atlanta, GA, USA. She is an IEEE Fellow for her contributes to “block transmissions over wireless fading channels.” Her research interests are in the areas of signal processing for communications and networks, signal estimation algorithms, coding theory, wireless communication theory, and sensor and ad hoc networks. Ma is a senior area editor for IEEE Signal Processing Letters and Elsevier Digital Signal Processing and has been an associate editor for the IEEE Transactions on Wireless Communications and Signal Processing Letters. She served as publication chair for the \emph{IEEE GLOBECOM} 2013, local arrangements chair for IEEE GlobeSIP 2014, and general chair for the ACM International Conference on Underwater Networks and Systems 2015 and 2019. Her recent research interests rely on intelligent wireless communication, decentralized networks, and sensor networks. 
\end{IEEEbiography}

\begin{IEEEbiography}[{\includegraphics[width=1in,height=1.25in,clip,keepaspectratio]{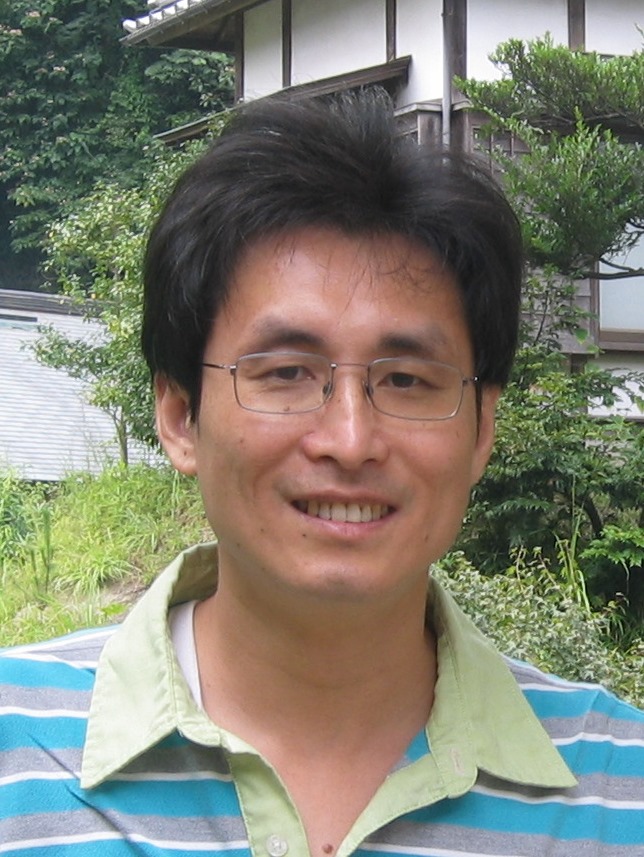}}]{Hsi-Sheng Goan} received his Ph.D. degrees in Physics from the University of Maryland, College Park, USA in 1999. He then worked as a postdoctoral research fellow at the University of Queensland, Brisbane, Australia from 1999-2001. From 2002-2004, he was a senior research fellow awarded the Hewlett-Packard Fellowship at the Center for Quantum Computer Technology at the University of New South Wales, Sydney, Australia before he took up a faculty position at the Department of Physics, National Taiwan University (NTU) in 2005. He is currently a Professor of Physics at NTU working in the fields of Quantum Computing and Quantum Information, Quantum Control, Mesoscopic (Nano) Physics, Quantum Optics, and Quantum Optomechanical and Electromechanical Systems.
Professor Goan has served as a member of the Editorial Boards of several international scientific journals, such as International Journal of Quantum Information, European Physical Journal: Quantum Technology, Chinese Journal of Physics, and Frontiers in ICT: Quantum Computing.
\end{IEEEbiography}

\EOD

\end{document}